\documentclass[11pt, letterpaper, logo, onecolumn, copyright, numbering]{acc}

\usepackage{url}

\usepackage{natbib}

\usepackage{amsthm}

\usepackage{times}

\definecolor{darkblue}{RGB}{32, 64, 129}
\definecolor{darkgreen}{RGB}{0, 110, 85}
\definecolor{darkred}{RGB}{153, 0, 0}
\definecolor{graytext}{gray}{0.45}

\definecolor{evaluatorcolor}{RGB}{255,230,230}    %
\definecolor{evaluatorframe}{RGB}{180,50,50}      %
\definecolor{taskgencolor}{RGB}{230,255,230}      %
\definecolor{taskgenframe}{RGB}{50,180,50}        %
\definecolor{executioncolor}{RGB}{230,230,255}    %
\definecolor{executionframe}{RGB}{50,50,180}      %
\definecolor{lightgray}{RGB}{240,240,240}
\definecolor{darkgray}{RGB}{80,80,80}

\definecolor{mintframe}{HTML}{7FB77E}
\definecolor{mintback}{HTML}{E9F6E9}    % lighter background

\definecolor{roseframe}{HTML}{E78BA2}  % darker version
\definecolor{roseback}{HTML}{FDE6EB}   % lighter version

\definecolor{skyframe}{HTML}{6CA6CD}  % darker blue tone
\definecolor{skyback}{HTML}{E6F2FA}   % lighter background

\newcommand{\method}{{PromptBridge }}

\usepackage{pifont}
\usepackage{longtable}

\usepackage{makecell}

\usepackage{amsmath}
\usepackage{pifont}

\usepackage{booktabs}
\usepackage{multirow}
\usepackage{graphicx}
\usepackage{float}
\usepackage{caption}
\usepackage{subcaption}
\usepackage{xspace}

\usepackage[ruled,vlined]{algorithm2e}

\usepackage[utf8]{inputenc}
\usepackage[T1]{fontenc}
\usepackage{xcolor}
\usepackage[most]{tcolorbox}
\usepackage{enumitem}
\usepackage{listings}

\definecolor{darkgray}{rgb}{0.3, 0.3, 0.3}
\definecolor{lightgray}{rgb}{0.95, 0.95, 0.95}
\definecolor{codegray}{rgb}{0.98, 0.98, 0.98}

\newtcolorbox{promptbox}[1]{
    colback=lightgray,
    colframe=darkgray,
    colbacktitle=darkgray,
    coltitle=white,
    boxrule=2pt,
    arc=0mm,
    left=10pt,
    right=10pt,
    top=10pt,
    bottom=10pt,
    fonttitle=\bfseries\large,
    title={#1},
    attach boxed title to top left={yshift=-2mm}
}

\usepackage{ragged2e}
\usepackage{wrapfig}
\usepackage{titletoc}

\usepackage[noend]{algpseudocode}

\definecolor{deepred}{rgb}{0.631,0.102,0.102}
\definecolor{skyblue}{HTML}{126da2}

\definecolor{accpurple}{HTML}{A100FF}
\hypersetup{
     colorlinks = true,
     breaklinks = true,
     linkcolor = accpurple,
     urlcolor = accpurple,
     citecolor = accpurple
     }

\definecolor{orange}{rgb}{1,0.5,0}

\algnewcommand{\LineComment}[1]{\State \(\triangleright\) #1}

\title{PromptBridge: Cross-Model Prompt Transfer for Large Language Models}

\author[1,2]{Yaxuan Wang}
  \author[2]{Quan Liu}
  \author[2]{Zhenting Wang}
  \author[2]{Zichao Li}
  \author[2]{Wei Wei}
  \author[1]{Yang Liu}
  \author[2]{Yujia Bao}

  \affil[1]{University of California, Santa Cruz}
  \affil[2]{Center for Advanced
   AI, Accenture}

  \date{\today}
  \correspondingauthor{Yujia Bao (\href{mailto:yujia.bao@accenture.com}{yujia.bao@accenture.com}).}

\begin{abstract}
Large language models (LLMs) underpin applications in code generation, mathematical reasoning, and agent-based workflows. In practice, systems access LLMs via commercial APIs or open-source deployments, and the model landscape (e.g., GPT, Claude, Llama) evolves rapidly. This rapid evolution forces frequent model switches driven by capability, cost, deployment constraints, and privacy. Yet prompts are highly model-sensitive: reusing a prompt engineered for one model on another often yields substantially worse performance than a prompt optimized for the target model. We term this phenomenon \emph{Model Drifting}. Through extensive empirical analysis across diverse LLM configurations, we show that model drifting is both common and severe.
To address this challenge, we introduce \emph{PromptBridge}, a training-free framework that preserves prompt effectiveness under model switches, enabling cross-model prompt transfer without costly per-task or per-model re-optimization. \emph{PromptBridge} requires only a small set of alignment tasks for calibration. It first applies \emph{Model-Adaptive Reflective Prompt Evolution (MAP-RPE)} to obtain task- and model-specific optimal prompts via iterative reflective refinement and quantitative evaluation. Using the resulting calibrated prompt pairs for the source and target models, \emph{PromptBridge} learns a cross-model prompt mapping. At test time, i.e., for an unseen task, given a source-model prompt, this mapping directly produces an optimized prompt for the target model.
Experiments in single-agent and multi-agent settings show that \emph{PromptBridge} consistently improves downstream accuracy while reducing migration effort. For example, when transferring from the source model to the target model, such as \texttt{o3}, \emph{PromptBridge} yields a 27.39\% improvement on \textsc{SWE-Bench} and a 39.44\% improvement on \textsc{Terminal-Bench} relative to direct transfer. These results establish cross-model prompt transferability as a critical requirement for sustainable LLM system development and demonstrate that \emph{PromptBridge} is an effective, practical, training-free solution for maintaining performance as models evolve. The code will be available at \href{https://github.com/supergirl-os/PromptBridge}{PromptBridge}.
\end{abstract}

\begin{document}

\maketitle
\def\Snospace~{Section }
\def\sectionautorefname{\Snospace}
\def\subsectionautorefname{\Snospace}
\def\subsubsectionautorefname{\Snospace}
\def\chapterautorefname{\Snospace}

\section{Introduction}
\label{sec:intro}

Large Language Models (LLMs) have demonstrated remarkable capabilities across diverse tasks, including code generation~\citep{jimenez2024swebench, jain2024livecodebench, dong2025survey, dong2025codescore}, mathematical reasoning~\citep{guan2025rstar, xia2025evaluating}, agent-based workflows~\citep{huang2023agentcoder,wang2024mixture,yuan2024evoagent, liu2025sew, zhang2025evoflow, wei2025browsecomp} and other complex problem-solving domains. LLM applications now range from single-turn question answering to complex, multi-agent systems~\citep{li2025beyond} that write code, analyze data, make decision~\citep{su2025learn} and call external tools~\citep{yao2023react, wu2024autogen, chen2023agentverse}.
In all cases, system behavior is governed by prompts that define the model’s role, guide its reasoning, and specify constraints and workflows. As applications mature, these systems accumulate carefully engineered prompt templates that encapsulate expert knowledge and are repeatedly reused across tasks.

New model releases (e.g., GPT~\citep{achiam2023gpt}, Gemini~\citep{comanici2025gemini}, Claude~\citep{anthropic2024introducing}, Llama~\citep{grattafiori2024llama}) routinely promise improvements in accuracy, latency, deployment, licensing, and privacy~\citep{regulation2018general,neel2023privacy}.
This churn raises a practical question:

\begin{center}
\textbf{Can prompts engineered for one LLM transfer optimally to another without re-tuning?}
\end{center}

\begin{figure}[t]
    \centering
    \includegraphics[width=1.0\textwidth]{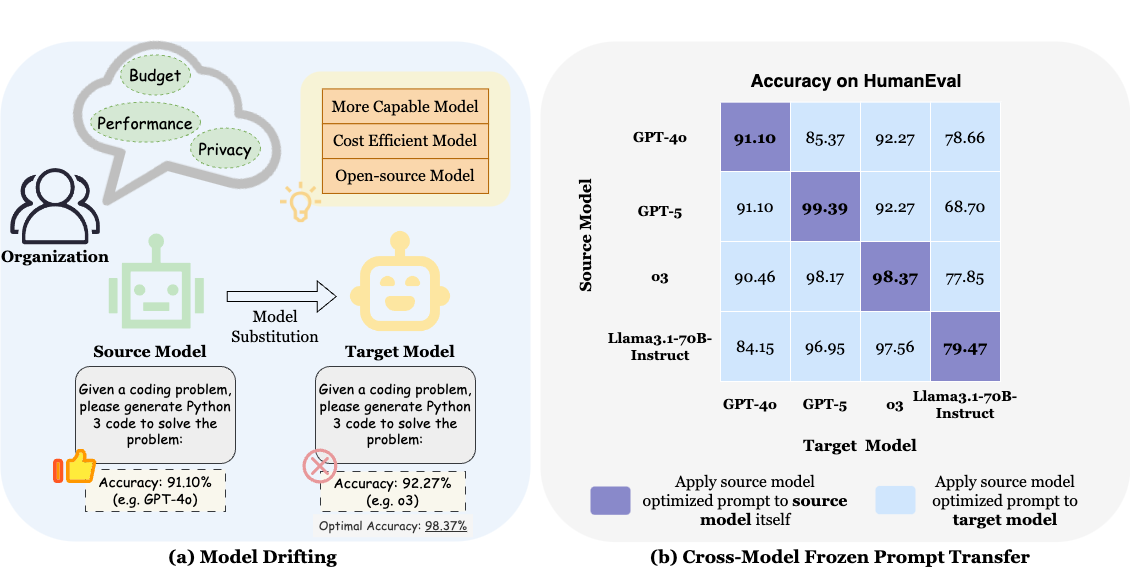}
    \vspace{-0.2cm}
    \caption{\textbf{Illustration of Model Drifting Problem.}
(a) Organizations may replace the source model with a more capable, cost-efficient, or open-source alternative. 
However, directly transferring the source-engineered prompt (e.g. \texttt{GPT-4o}) to a stronger target model (e.g. \texttt{o3}) yields only 92.27\% 
accuracy on the HumanEval dataset, far below the target model’s achievable 98.37\%, revealing that prompts engineered or optimized for one model do not reliably generalize to another.
(b) Cross-model evaluation on HumanEval demonstrates model drifting: prompts designed for a source model frequently degrade when applied to different target models. 
    }\label{fig:pre}
    \vspace{-0.4cm}
\end{figure}

Empirically, the answer is often no.
Here we use transfer-learning terminology: the \emph{source model} is the model on which prompts are tuned, and the \emph{target model} is the new model to which those prompts are applied.
Prompts engineered or optimized for a source model frequently underperform when applied to a target model, a phenomenon we term \textbf{Model Difting}.
\autoref{fig:pre} illustrates the pattern.
Panel (a) shows the common scenario: an organization replaces its source model (\texttt{GPT-4o}~\citep{openai_gpt4o_system_card_2024}) with a more capable alternative (\texttt{o3}~\citep{openai_o3_o4mini_system_card_2025}), yet the source-optimal prompt no longer remains optimal for the target.
Panel (b) quantifies drift on HumanEval~\citep{chen2021evaluating}: the best prompt for \texttt{GPT-5}~\citep{openai_gpt5_system_card_2025} reaches 99.39\% on \texttt{GPT-5} but yields 68.70\% when transferred to \texttt{Llama-3.1-70B-Instruct}~\citep{grattafiori2024llama}, while \texttt{Llama-3.1-70B-Instruct}'s own optimal prompt achieves 79.47\% on its own. The drift from \texttt{GPT-5} to \texttt{Llama-3.1-70B-Instruct} is therefore 68.70\% vs. 79.47\%. Conversely, the \texttt{Llama-3.1-70B-Instruct} optimal prompt attains 96.95\% on \texttt{GPT-5}, below \texttt{GPT-5}'s optimum of 99.39\% (a 2.44-point drift).
Additional analysis about this phenomenon appears in~\autoref{app:pre:necessity}.
Such transfer gaps occur across model families and across coding, agentic, and planning tasks, creating a barrier to model migration and A/B testing without re-engineering prompt libraries.

Why does model drift arise?
Vendors train and align models with different corpora, tokenization schemes, role tags, and human-feedback criteria.
For instance, Llama~3~\citep{grattafiori2024llama} introduces an \texttt{ipython} role for tool calls—an instruction absent from GPT models.
Models trained by different companies (e.g., Qwen~\citep{yang2025qwen3}, DeepSeek~\citep{guo2025deepseek} vs. Mistral~\citep{jiang2023mistral7b}) ingest different linguistic and domain distributions.
These alignment and interface mismatches make a prompt fit one model but mis-specify another, forcing organizations to rewrite and re-validate prompts when swapping LLMs.

This work tackles the prompt-transfer problem at two levels:

(1) \textbf{Quantifying Model Drifting.}
We formalize drift as the transfer gap to the target model's own optimal prompt.
For a task $T$, source model $M_s$, target model $M_t$, and performance metric $A(\cdot)$, let $p^*_{M,T}$ denote the model-preferred prompt that maximizes $A(M,T,p)$.
We define the transfer gap from $M_s$ to $M_t$ on $T$ as
$\Delta(M_s \!\rightarrow\! M_t, T) = A(M_t, T, p^*_{M_s,T}) - A(M_t, T, p^*_{M_t,T})$,
i.e., the shortfall of the transferred source-optimal prompt relative to the target model’s own optimum.
We empirically characterize this gap across models and tasks, revealing systematic drift even within model families.

(2) \textbf{PromptBridge: training-free cross-model prompt adaptation.}
Rather than re-optimizing prompts anew for every model and task, PromptBridge uses a small calibration suite of \emph{alignment tasks} for which we obtain source- and target-optimal prompts.
Given an unseen task and a source prompt, PromptBridge maps the source prompt to a target-style prompt by leveraging these calibration anchors, enabling zero-shot, in-context prompt adaptation without access to evaluation data for the unseen task.
To obtain high-quality anchors, we introduce \textbf{Model-Adaptive Reflective Prompt Evolution (MAP-RPE)}, an evaluation-guided, reflection-driven, island-based prompt search that produces model-preferred, task-specific prompts.
MAP-RPE supports both our drift measurement (by approximating $p^*_{M,T}$) and PromptBridge’s calibration.

We evaluate PromptBridge across seven LLMs and eight benchmarks under single-agent and multi-agent settings, with comprehensive ablations and preliminary analysis of model drifting.
PromptBridge consistently reduces the transfer gap and maintains or improves performance while substantially lowering migration overhead across different models.

\textbf{Contributions.}
Our main contributions are summarized as follows:
\begin{itemize}
\item We introduce and empirically quantify the problem of \textbf{Model Drifting}, which arises when the underlying LLM in a system is replaced or updated, leading to performance shifts under the same prompt. We further establish the task of prompt transfer under model drifting and provide a general framework for addressing it.
\item We propose a \textbf{Model-Adaptive Reflective Prompt Evolution (MAP-RPE)} method that calibrates prompts for a given specific model through reflective, metric-driven optimization, producing both task- and model-specific optimized prompts and serving the drift measurement process.
\item We develop \textbf{PromptBridge}, a training-free cross-model prompt transfer framework that learns transferable prompt transformation mappings from alignment tasks and enables zero-shot prompt adaptation to unseen tasks across different target models.
\item Extensive experiments across single-agent and multi-agent settings demonstrate the effectiveness of our framework and underscore the importance of the model drifting problem.
\end{itemize}
\section{Related Work}
\label{sec:related}

\paragraph{Prompt Optimization }
A closely related line of research investigates how to optimize and adapt prompts to better elicit desirable behaviors from LLMs. Foundational strategies such as Chain-of-Thought prompting~\citep{wei2022chain} substantially enhance reasoning abilities across diverse domains. Subsequent works explore automated prompt optimization using advanced online LLMs to iteratively refine prompts~\citep{zhou2022large, cheng2023black, yang2023large, agarwal2024promptwizard}.
\citet{fernando2023promptbreeder} introduce PromptBreeder, which evolves a population of task prompts through mutation and fitness evaluation on a validation set. MultiPrompter~\citep{kim2023multiprompter} formulates prompt optimization as a cooperative game among multiple prompters that sequentially refine the same prompt. MARS~\citep{zhang2025mars} adopts a multi-agent architecture that generates and evaluates candidate prompts via diverse search strategies.
Beyond direct LLM-based optimization, evolutionary algorithms have also been applied to improve LLM performance. EvoPrompt~\citep{guo2023connecting} connects LLMs with evolutionary search to progressively populate optimized prompts, though it lacks integration with feedback. AlphaEvolve~\citep{novikov2025alphaevolve} and its open-source variant OpenEvolve~\citep{openevolve} demonstrate that evolutionary rewriting within prompts can yield substantial performance gains in code generation tasks. GEPA~\citep{agrawal2025gepa} employs reflective feedback and Pareto-aware optimization to balance competing objectives when evolving prompt candidates.
Despite the effectiveness of previous prompt optimization method, these methods often rely on large, proprietary LLMs and show limited transferability to smaller models~\citep{wang2023promptagent}. To address this, \citet{zhu2025rethinking} propose several design principles for constructing more generalizable prompts. The importance of prompt design that generalizes across model families remains underexplored. \citet{rakotonirina2023can} partially address this by inducing prompts via model mixing during training, producing prompts that generalize better across architectures.

\paragraph{Prompt Sensitivity } Previous study has demonstrated that LLM is sensitive to prompt~\citep{wei2025paft, zhu2025rethinking}.
Minor deviations between user-provided instructions and the training instructions can result  in significant performance degradation~\citep{wei2025paft}. The existing research primarily emphasizes the importance of adapting prompts to specific tasks~\citep{zheng2024decomposed, wang2023multitask}, or specific modality~\citep{zhang2025release}, rather than specific LLMs. MAPO~\citep{chen2024mapo} first demonstrate that different prompts should be adapted to different LLMs to enhance their capabilities across various downstream tasks. The LLM agent might also suffer from prompt sensitivity~\citep{verma2024brittle}, where simple modifications in the prompt can already exert significant but unexpected degradation of performance~\citep{zhou2023batch,liu2024self}. 
However, these works typically assume a fixed model. In contrast, the model drifting problem studied in this paper concerns how the optimal prompt itself evolves as the model architecture, or alignment procedure change over time or across model families. This phenomenon extends beyond prompt sensitivity: rather than a local perturbation effect, model drifting captures a systematic semantic shift in how different or updated LLMs interpret the same instruction, necessitating adaptive calibration and transfer methods to maintain performance consistency.

\paragraph{Optimizing AI systems and Agents }
Optimizing AI systems composed of multiple interacting LLM modules remains a fundamental challenge due to the complex dependencies among components. DSPy~\citep{khattab2024dspy} automates exemplar and instruction design to enhance the performance of multi-module LLM pipelines. TextGrad~\citep{yuksekgonul2025optimizing} introduces a differentiable framework that backpropagates textual feedback across modules, allowing downstream performance signals to refine upstream prompts. MIPROv2~\citep{opsahl2024optimizing} jointly aligns instructions and few-shot examples via Bayesian optimization. Optimas~\citep{wu2025optimas} further improves multi-module systems by optimizing locally aligned rewards that are globally coordinated across the entire architecture.
In multi-agent settings, MASS~\citep{zhou2025multi} serves as a plug-and-play prompt and workflow optimizer that mitigates prompt sensitivity arising from inter-agent compounding effects. Together, these studies highlight the growing interest in system-level optimization methods that move beyond individual prompts or models to holistically improve multi-component AI systems.

\paragraph{Prompt Transfer } Research on prompt transfer investigates how to generalize and reuse effective prompts across tasks, modalities, or models. Existing work focuses mainly on cross-task transfer~\citep{zheng2024decomposed, wang2023multitask}, while \citet{zhang2025release} explore cross-modality transfer, demonstrating that prompts designed for one modality can partially transfer to another.
\citet{su2022transferability} investigate the cross-model transferability of soft prompts by training a prompt projector through prompt tuning on a specific task. However, their findings are limited to soft prompt and pretrained language models such as RoBERTa, without exploring broader model families or more recent LLM architectures.
\citet{chen2024mapo} introduce MAPO, a model-adaptive prompt optimizer that learns to tailor prompts for specific LLMs, showing that prompt effectiveness strongly depends on the underlying model family. \citet{wang2025efficient} propose a privacy-preserving soft prompt transfer framework, which trains soft prompts on smaller models and securely transfers them to larger ones, thereby improving transferability while maintaining privacy guarantees.
Despite these advances, cross model prompt transfer remains underexplored for scenarios involving model drifting or complex LLM systems, where prompt adaptability and consistency are critical yet challenging to maintain.

\section{Problem Formulation}

We focus on prompt template transfer when switching from one LLM to another, which fall within the task- and model-adaptive prompt optimization area. The definition of Model Drifting is given below:

\textit{\textbf{Model Drifting.} Given an LLM-based system $\Phi_{M,p}$, where $M$ denotes the underlying LLM and $p$ represents the prompt template that guide model behavior, we may wish to replace the source model $M_{s}$ with a target model $M_{t}$ in practice due to budget constraints, deployment requirements, or special demands. Such a replacement inevitably alters the system’s behavior and performance characteristics. We refer to the performance drift induced by the model substitution as \emph{model drifting}.}

Formally, let $A(\cdot)$ denote a performance metric of an LLM-based system. When the underlying model is replaced, the same prompt that was previously optimal for the source model often yields suboptimal performance on the target model for task $T$:
\begin{equation}
    \underbrace{A\!\left(M_t, T, p^*_{M_s, T}\right)}_{\text{after model switch}}
<
\underbrace{\max_{ p_{M_t, T}} A\!\left(M_t, T, p_{M_t, T}\right)}_{\text{target-optimal prompt performance}}
\end{equation}

The core challenge is that the prompt $p^*_{M_s,T}$, which was optimal for $M_s$, is no longer optimal for $M_t$. Therefore, effective prompt recalibration or transfer is required to restore alignment between the prompt and the target model’s behavioral distribution.

In this work, we study model drifting under two system configurations: the single-agent setting and the multi-agent setting, with our primary focus on the single-agent case.
The single-agent setting also fully encompasses traditional single-LLM tasks: one model is responsible for the entire problem, and model drifting simply means swapping the underlying LLM while keeping the overall task and agent structure fixed.
In the multi-agent setting, model drifting can occur at two levels: Local model drifting refers to updating the source LLM of a specific agent to a new target model, while other agents remain unchanged; Global model drifting refers to simultaneously replacing the source models of all agents with their corresponding target models.

\paragraph{Prompt Transfer under Model Drifting.} 
Let $S = \{S_1, S_2, \dots, S_m\}$ denote the set of \emph{alignment tasks} used as model-adaptive prompt alignment data, and $T = \{T_1, T_2, \dots, T_q\}$ denote the set of \emph{unseen tasks}. 
For each alignment task $S_i \in S$, let $p^*_{M_s, S_i}$ and  $p^{*}_{M_t, S_i}$  represent the task-specific optimal prompts for the source model $M_s$ and target model $M_t$, respectively. 
Similarly, for each unseen task $T_j \in T$, let $p^{*}_{M_s, T_j}$ and $p^{*}_{M_t, T_j}$ denote the corresponding optimal prompts.
Given the unseen tasks $T$, prompt transfer aims to learn a mapping: $
\mathcal{T}_{M_s \rightarrow M_t;  S}:\;
p_{M_s, T}
\;\longmapsto\;
p^*_{M_t, T}
$, 
where the mapping captures systematic relationships between model- and task-specific prompt structures.
The transfer function $\mathcal{T}$ is learned from the alignment tasks $S$, for which both the source-model prompts and the corresponding target-model prompts are either known or can be obtained through calibration methods. Once learned, this transformation is applied to unseen tasks, where only the source-model prompt $p_{M_s, T}$ is available. 
Applying $\mathcal{T}$ produces an adapted target-model prompt $\widehat{p}_{M_t, T}$ that aims to (i) match or surpass the performance of the source-model prompt $A\!\left(M_s, T, p_{M_s, T}\right)$, and (ii) approximate as closely as possible the performance of the true task-optimal target-model prompt $A\!\left(M_t, T, p^{*}_{M_t,T}\right)$.

\section{Method}
\label{sec:method}
\begin{figure}[t]
    \centering
    \includegraphics[width=\textwidth]{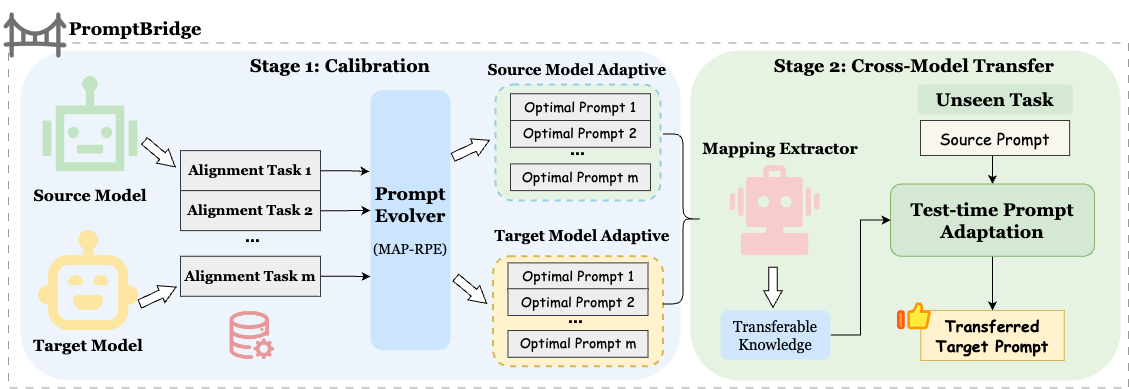}
    % \vspace{-15pt}
    \caption{\textbf{Overview of the proposed PromptBridge framework.} 
    The framework operates in two stages: 1) Calibration, where Model-Adaptive Reflective Prompt Evolution (MAP-RPE) optimizes task-specific prompts for each model using quantitative feedback; and 2) Cross-Model Transfer, where pairs of calibrated source–target prompts are used to learn a transferable prompt mapping function $\mathcal{T}$. This learned mapping acts as a bridge that enables the system to synthesize target-compatible prompts for unseen tasks, supporting zero-shot prompt transfer without additional training.}
    \label{fig:method}
\end{figure}

In this section, we first introduce \textbf{PromptBridge}, a model-adaptive prompt transfer framework that leverages calibrated prompts on alignment tasks to learn transferable prompt knowledge across different LLMs. We then present our Model-Adaptive Reflective Prompt Evolution (MAP-RPE) method, which optimizes prompts for a specific model in a reflective and adaptive manner. MAP-RPE is used as the calibration mechanism to obtain optimal prompts on the alignment tasks, which subsequently support the model-adaptive transfer process. \autoref{fig:method} provides an overview of the proposed approach.

\subsection{Cross-Model Prompt Transfer}
\paragraph{Transfer Mapping Learning.}
Let $S = \{S_1, S_2, \dots, S_m\}$ denote the set of alignment tasks used for learning model-adaptive prompt alignment. 
For each task $S_i$, we first obtain the task-specific optimal prompts using MAP-RPE, denoted as $\textrm{Evolver}(\cdot)$:  
\[
p^*_{M_s, S_i}= \textrm{Evolver}(p_{M_s, S_i}),  \qquad
p^*_{M_t, S_i}= \textrm{Evolver}(p_{M_t, S_i})
\]
The objective of the transfer module is to learn the mapping:
\begin{equation}
\mathcal{T}_{M_s \rightarrow M_t;  S}:\;
p^*_{M_s, S_i}
\;\longmapsto\;
p^*_{M_t, S_i}
% \mathcal{T}_{\mathcal{M}_{\textrm{source}} \rightarrow \mathcal{M}_{\textrm{target}};S}:
% \mathcal{P}^{*}_{\textrm{source}}(S_i)
% \;\longmapsto\;
% \mathcal{P}^{*}_{\textrm{target}}(S_i),
\qquad S_i \in S.
\end{equation}
which captures how prompts should be systematically reformulated when moving from the source model to the target model.  
To extract the transferable knowledge, we employ a \emph{Mapping Extractor} (e.g., a high-capability LLM) to analyze each pair $(p^*_{M_s, S_i}, p^*_{M_t, S_i})$  and generate a concise description of the model-specific transformation pattern. By leveraging the LLM's ability to summarize structural and stylistic differences, the mapping-model distills the consistent prompt-level adjustments required when transferring prompts from the source model to the target model. This distilled summary constitutes the transferable knowledge, describing how task-optimal prompts for the source model should be reformulated to align with the behavioral, stylistic, and reasoning characteristics of the target model.

\begin{figure}[t]
    \centering
    \includegraphics[width=1.0\textwidth]{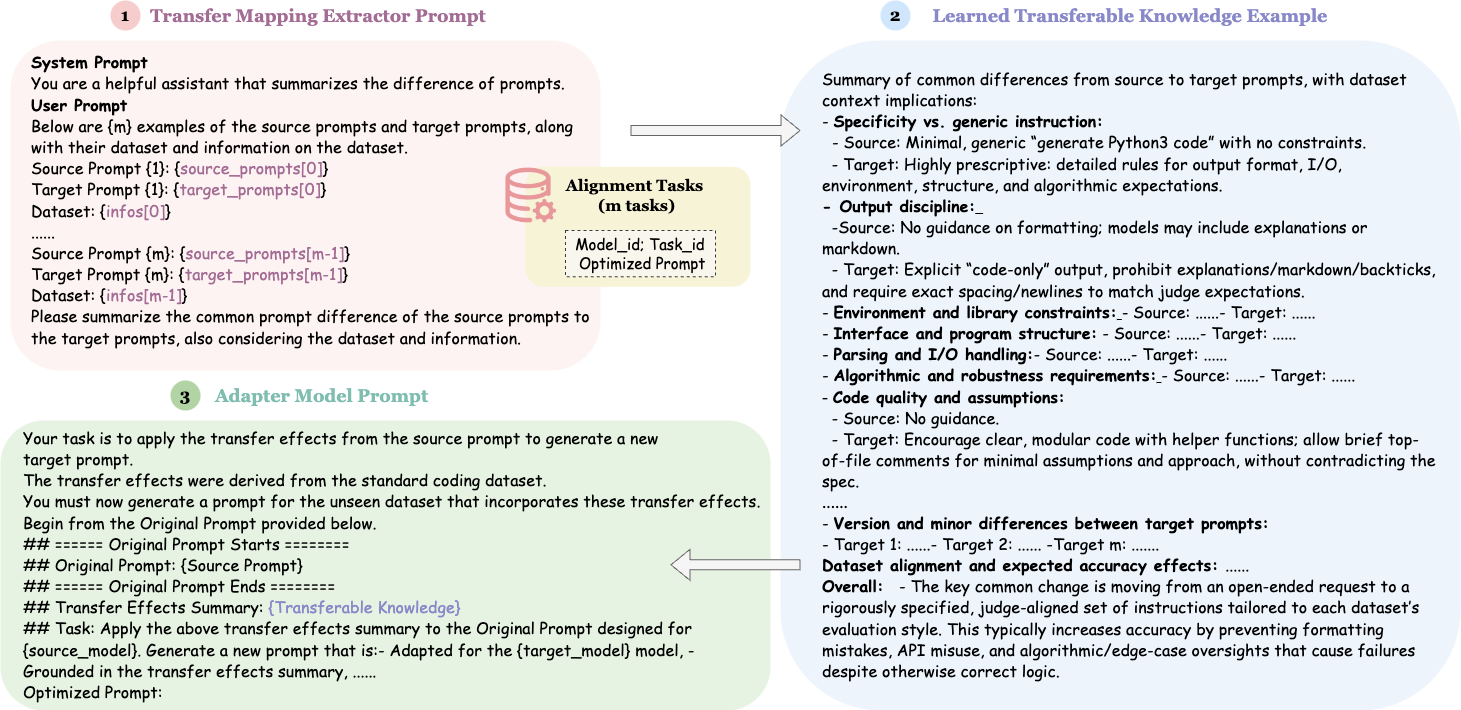}
    \vspace{-0.2cm}
    \caption{\textbf{A demonstration of PromptBridge.}
1) shows the prompt template used by the Mapping Extractor for learning transferable mappings between source and target prompts.
2) presents an example of the learned transferable knowledge, which may contain multiple sections analyzing cross-model prompt differences.
3) illustrates the prompt template used by the Adapter Model, which applies the learned mapping to adapt prompts for unseen tasks.
    }\label{fig:example}
    \vspace{-0.4cm}
\end{figure}

\paragraph{Test-Time Prompt Adaptation.}
For an unseen task $T_j \in T$ with its source prompt 
$p_{M_s, T_j}$, 
PromptBridge applies the learned mapping in context to synthesize a target-model-compatible prompt:
\begin{equation}
\widehat{P}_{M_t, T_j}
= 
\mathcal{T}_{M_s \rightarrow M_t;  S}
\big( p_{M_s, T_j}\big) \;\approx\;
p^*_{M_t, T_j},
\qquad T_j \in T.
\end{equation}
where $\mathcal{T}$ denotes the transformation extracted from the alignment tasks. 
At test time, this transformation is implemented through an \emph{adapter model} that conditions both the source prompt and the learned structural mapping. The adapter predicts the structural edits, formatting adjustments, and semantic refinements required for compatibility with the target model, and then generates a fully adapted prompt $\widehat{P}_{M_t, T_j}$.
This design enables efficient, zero-shot prompt adaptation to new tasks and target models without additional training or calibration, allowing our approach to generalize beyond the alignment tasks used to learn $\mathcal{T}$. 

\textbf{PromptBridge} introduces a new paradigm for cross-model prompt generalization. It 1) formulates prompt transfer under model drifting as a learnable transformation problem rather than performing independent prompt re-optimization or manually human design, allowing an optimized prompt from one model to be systematically transformed into an effective prompt for another. Details on the feasibility of cross-model prompt transfer are presented in~\autoref{app:pre:feasibility} in the Appendix; 2) extracts model-level differences through linguistic summarization instead of relying on heavy gradient-based signals; and 3) enables zero-shot adaptation to different unseen tasks utilizing the mapping learned from alignment tasks without re-optimization. 
Together, these features make PromptBridge a scalable, effective, and low-cost bridge for aligning prompt behavior across continuously evolving LLM architectures. \autoref{fig:example} illustrates how PromptBridge learns transferable mappings with the Mapping Extractor and applies them through the Adapter Model to adapt prompts for unseen tasks.
The complete prompt template used for this process is provided in~\autoref{app:prompt:method} in the Appendix.

\subsection{Calibration: Model-Adaptive Reflective Prompt Evolution}

Inspired by AlphaEvolve~\citep{novikov2025alphaevolve}, we propose \textbf{Model-Adaptive Reflective Prompt Evolution (MAP-RPE)}, a calibration method that enables a specific model to iteratively evolve and align its prompt through reflective feedback, behavior-aware scoring, and island-based population evolution.
Unlike AlphaEvolve, which directly evolves program outputs through population-based code mutation and selection, MAP-RPE focuses on evolving the prompt template itself. This shift allows the model to adaptively improve its own prompt without modifying internal parameters or accessing gradients. The framework leverages a \emph{reflection model} to analyze past generations and propose refined prompts, while the \emph{specific model} executes these prompts to generate responses that provide behavioral and performance feedback.

\paragraph{Reflective Prompt Optimization Process.}
Instead of typical generate-and-select paradigm~\citep{wang2023promptagent}, our method begins with the source prompt $p_{M_s, S_i}$ and iteratively refines it for the specific model. For each iteration, a batch of randomly selected inputs from the current task $S_i$ is used to query the target model, producing responses that are then evaluated using designed quantitative metrics, such as performance score and behavior score, to capture both task accuracy and response quality. The reflection model aggregates these results, identifies weaknesses or inconsistencies, and proposes an improved prompt. Through repeated reflection and feedback, the prompt gradually converges to an optimized version $p^*_{M, S_i}$, where $M$ denotes the specific model.

\paragraph{Island-Based Evolution for Diversity.}
To prevent premature convergence to local optima, MAP-RPE maintains an island-based population of diverse prompts. Each island stores variants that capture different behavioral patterns or task-solving strategies. In each iteration, the current best-performing prompt serves as the parent to generate new candidate prompts inspired by accumulated feedback. These candidates are re-evaluated and reflected, allowing the system to continuously balance exploration and exploitation in the prompt space.

\paragraph{Compared with GEPA}
Unlike GEPA~\citep{agrawal2025gepa}, which relies primarily on textual feedback and Pareto-aware optimization over prompt candidates, our method adopts a model-adaptive and metric-driven optimization paradigm. It introduces explicit quantitative metrics, including performance, behavioral, and task-specific scores to evaluate how well each prompt aligns with the target model’s performance characteristics. Guided by these evaluations, the optimizer iteratively refines the source prompt into an optimal variant $p^*_{M_t, S_i}$ tailored to the specific target model. By maintaining diverse islands of prompt candidates and addressing the prompt–model mismatch problem common in cross-model transfer, MAP-RPE achieves more structured, efficient, and adaptive prompt calibration than GEPA’s general reflection-based framework.

Overall, MAP-RPE provides a gradient-free, model-adaptive calibration mechanism that evolves prompts through structured reflection and population diversity, enabling each specific model to achieve its optimal task-specific prompt.~\autoref{alg:prompt-evolving} shows the reflective loop of MAP-RPE. 
\section{Experiments}

\begin{table*}
\centering
\caption{Pass@1 Accuracy on several code generation datasets when switching the original model \texttt{GPT-4o} to the target models, including \texttt{o3}, \texttt{o4-mini}, \texttt{Llama3.1-70B-Instruct}. Direct Transfer refers to apply the source prompt to the target model directly. The higher the accuracy the better. The best transfer results are highlighted in \textbf{Bold}.
} 
\resizebox{0.9\textwidth}{!}{
\begin{tabular}{c|ccccc}
\toprule
 \multicolumn{1}{c}{\textbf{Method}} & \multicolumn{1}{c}{\textbf{HumanEval}}  & \multicolumn{1}{c}{\textbf{MBPP}} & \multicolumn{1}{c}{\textbf{APPS}}    & \multicolumn{1}{c}{\textbf{xCodeEval}} & \multicolumn{1}{c}{\textbf{CodeContests}}\\ 
\midrule
Source Model: GPT-4o & 91.10 & 79.80 & 12.00 & 37.03 & 5.45 \\
\midrule
\multicolumn{6}{c}{\textbf{Target Model: o3}} \\
\midrule
Direct Transfer & 92.27 & 77.92 & 32.67  & 66.04 & 48.61\\
GPT-5 Optimizer & 92.68 & 79.93 & \textbf{36.44} & 72.64 & 31.88 \\
MIPROv2~\citep{opsahl2024optimizing} & 93.70 & 63.81 & 32.67 & 65.09 & 45.65 \\
GEPA~\citep{agrawal2025gepa} & 96.95 & 80.01 & 32.89 & 69.49 & 46.19 \\
\rowcolor{gray!30}  
\method (Ours) &\textbf{97.15} & \textbf{80.44} & \textbf{36.44} & \textbf{74.84} & \textbf{56.36} \\
\midrule
\multicolumn{6}{c}{\textbf{Target Model: o4-mini}} \\
\midrule
Direct Transfer & 96.54 & 79.09 & 37.78 & 72.01 & 40.12 \\
GPT-5 Optimizer & 98.17& 70.03 & \textbf{38.44} & 74.53 & 57.70  \\
MIPROv2~\citep{opsahl2024optimizing} & 96.95 & 77.92 & 35.33 & 74.84 & 48.28\\
GEPA~\citep{agrawal2025gepa} & 97.56 & 76.49 & 37.33 & \textbf{81.13} & 55.15\\
\rowcolor{gray!30}  \method (Ours) & \textbf{98.37} & \textbf{80.60} & 38.00 & 77.67 & \textbf{58.79} \\
\midrule
\multicolumn{6}{c}{\textbf{Target Model: Llama3.1-70B-Instruct}} \\
\midrule
Direct Transfer & 68.70 & 65.57 & 10.00 & 21.38 & 20.20\\
GPT-5 Optimizer & 78.66 & 73.38 & 7.78& 23.58 & 19.80 \\
MIPROv2~\citep{opsahl2024optimizing} & 78.25 & 68.26 & \textbf{11.33} & 18.87 & 21.82 \\
GEPA~\citep{agrawal2025gepa} & 35.98 & 25.86 & 10.22 & 22.01 &18.99 \\
\rowcolor{gray!30}  \method (Ours) & \textbf{79.88} & \textbf{73.64} & 11.11 & \textbf{24.53} & \textbf{23.84}\\
\bottomrule
\end{tabular}
}

\label{tab:gpt4o-main-code}
\end{table*}

We conduct extensive experiments to demonstrate the effectiveness of the proposed prompt transfer method (\S~\ref{exp:prompt_transfer}) and prompt calibration methods (\S~\ref{exp:prompt_calibration}) in single agent and multi-agent settings. Our experiments cover all three real-world application cases: transferring to a more advanced model, a more cost-efficient model, and an open-source model, demonstrating the effectiveness and generalization of the proposed cross-model transfer method.

\subsection{Alignment and Unseen Tasks.} 

\textbf{Alignment Tasks.}
Alignment tasks refer to the set of training tasks used to calibrate and align the prompts of the source and target models. These tasks provide the foundation for learning the transferable prompt mapping in our method. We use synthetic-code-generations~\citep{wei2023magicoder} and the train split of CodeContests~\citep{li2022competition} as the alignment tasks, which includes basic code generation and complex competitive coding problems.

\paragraph{Unseen Tasks.} For extensive evaluation, we use five coding benchmark datasets: two from basic programming and three from complex competitive programming domains. Two simple coding generation benchmarks are: HumanEval~\citep{chen2021evaluating} and MBPP~\citep{austin2021program}. Due to the absence of sample I/O in MBPP, the evaluation involves randomly removing one test-case from MBPP-ET~\citep{dong2025codescore} for each problem and provide this test-case as a sample I/O for the problem following~\citep{islam2024mapcoder}. 
Three complex competitive benchmarks are: Automated Programming Progress Standard (APPS)~\citep{hendrycks2021measuring}, xCodeEval~\citep{khan2023xcodeeval}, and CodeContests~\citep{li2022competition}.

Moreover, we use SWE-bench Verified~\citep{jimenez2024swebench} that evaluates how well LLM agents automatically solve the coding task. Given a code repository and a task instruction, the agent is expected to make changes to the repository in order to fulfill the task. The Verified subset is a human-filtered subset of 500 instances. Terminal-Bench~\citep{tbench_2025} measures the model’s ability to accomplish complex tasks in a terminal environment. 
We also evaluate on the validation split of TravelPlanner~\citep{xie2024travelplanner}. In the two-stage mode, it uses the ReAct framework for information collection and it allows to assess how different LLMs perform under a uniform tool-use framework. The
agents are required to give the plan directly based on the
information collected by themselves, without employing
any other planning strategies.  In the sole-planning mode, it aims to assess if the strategies proven effective in other planning benchmarks maintain their efficacy in TravelPlanner.

\subsection{Experimental Setup}

\paragraph{Baselines.}  
We evaluate three training-free prompt transfer baselines: 1) GPT-5 Optimizer, a powerful prompt optimizer that improving existing prompts and migrating prompts for GPT-5 and other OpenAI models; 2) MIPROv2~\citep{opsahl2024optimizing}, a widely used prompt optimizer and has been integrated into the DSPy~\citep{khattab2024dspy} framework. All MIPROv2 optimization runs are performer with the $auto=heavy$ setting. 3) GEPA~\citep{agrawal2025gepa}, a prompt optimizer that thoroughly incorporates natural language reflection to learn high-level rules from trial and error. Prompt optimizer with star (MIPROv2*, GEPA*) refer to directly optimized on the test benchmark itself with randomly select 50 questions from it. 
The detailed experiment setup introduction can be found in the \autoref{app:setting}. The exact optimized prompt templates are provided in the \autoref{app:prompt_template}.

\paragraph{Implementation Details.} 
We use \texttt{GPT-4o}~\citep{openai_gpt4o_system_card_2024} as the soucre model, \texttt{o3}~\citep{openai_o3_o4mini_system_card_2025}, \texttt{o4-mini}~\citep{openai_o3_o4mini_system_card_2025}, \texttt{Llama3.1-70B-Instruct}~\citep{grattafiori2024llama}, \texttt{Qwen3-32B}~\citep{yang2025qwen3} and \texttt{Gemma3-27B-it}~\citep{team2025gemma} as the target models for coding benchmarks. We utilize \texttt{GPT-5}~\citep{openai_gpt5_system_card_2025} as the Mapping Extractor and Adapter Model to transfer prompt. 
% When calling API, we set the temperature as 0.0 to reduce the randomness of each experiments. For each experiment, we run three times and calculate the average evaluation results.
For SWE-Bench, we utilize the framework of mini-SWE-agent~\citep{yang2024sweagent}. We use the Terminus framework for evaluation on Terminal-Bench following~\citet{zeng2025glm}.
We utilize MapCoder~\citep{islam2024mapcoder} as our multi agent system framework. It consists of four LLM agents specifically designed to handle coding problems: recalling relevant examples (Retrieval Agent), planing (Planning Agent), code generation (Coding Agent) and debugging (Debugging Agent).

\paragraph{Evaluation Metric and Setup.} We report average Pass@1 accuracy for all coding benchmarks, considering a model successful if its first generated solution is correct. When calling API, we set the temperature as 0.0 to reduce the randomness of each experiments. For \texttt{o3} and \texttt{o4-mini}, the Azure deployed models do not allow setting temperature manually, so we use the default setting to evaluate on benchmarks, and to reduce the randomness, we run each experiment several times and report the average result. For \texttt{Qwen3-32B} model, on APPS and CodeConetest, we set the max tokens to be 10000. For all datasets, we use temperature = 0.7. 
For each experiment, we run three times and report the average accuracy to ensure stability and reproducibility of the results.
For the SWE-Bench Verified benchmark, we instead report the resolved rate. For Terminal-Bench, we report the accuracy.

\begin{figure}[t]
    \centering
    \includegraphics[width=0.85\textwidth]{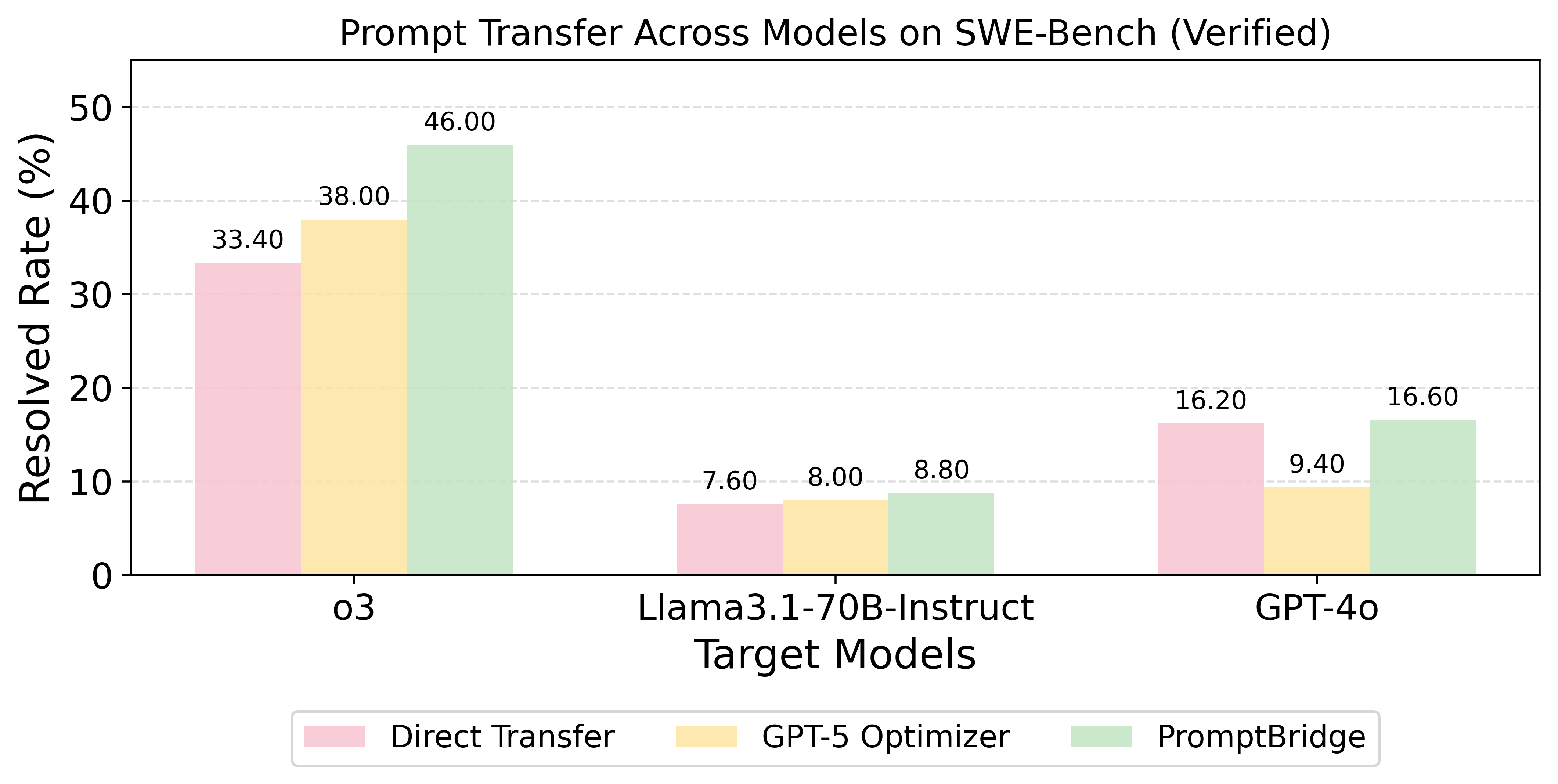}
    % \vspace{-15pt}
    \caption{\textbf{Results on SWE-Bench Verified.} We use \texttt{o4-mini} as the source model and evaluate transferability on \texttt{o3}, \texttt{Llama-3.1-70B-Instruct}, and \texttt{GPT-4o} as target models. On the source, the resolved rate with the default prompt is 38.60\%, which increases to 42.20\% after calibration with the optimized prompt.
    Direct Transfer refers to directly applying the optimized prompt from \texttt{o4-mini} to the other models without further adaptation.}
    \label{fig:swe-results}
\end{figure}

% Optimized represents the results obtained after refining this prompt for o4-mini to enhance its performance.
\begin{figure}[t]
    \centering
    \includegraphics[width=0.85\textwidth]{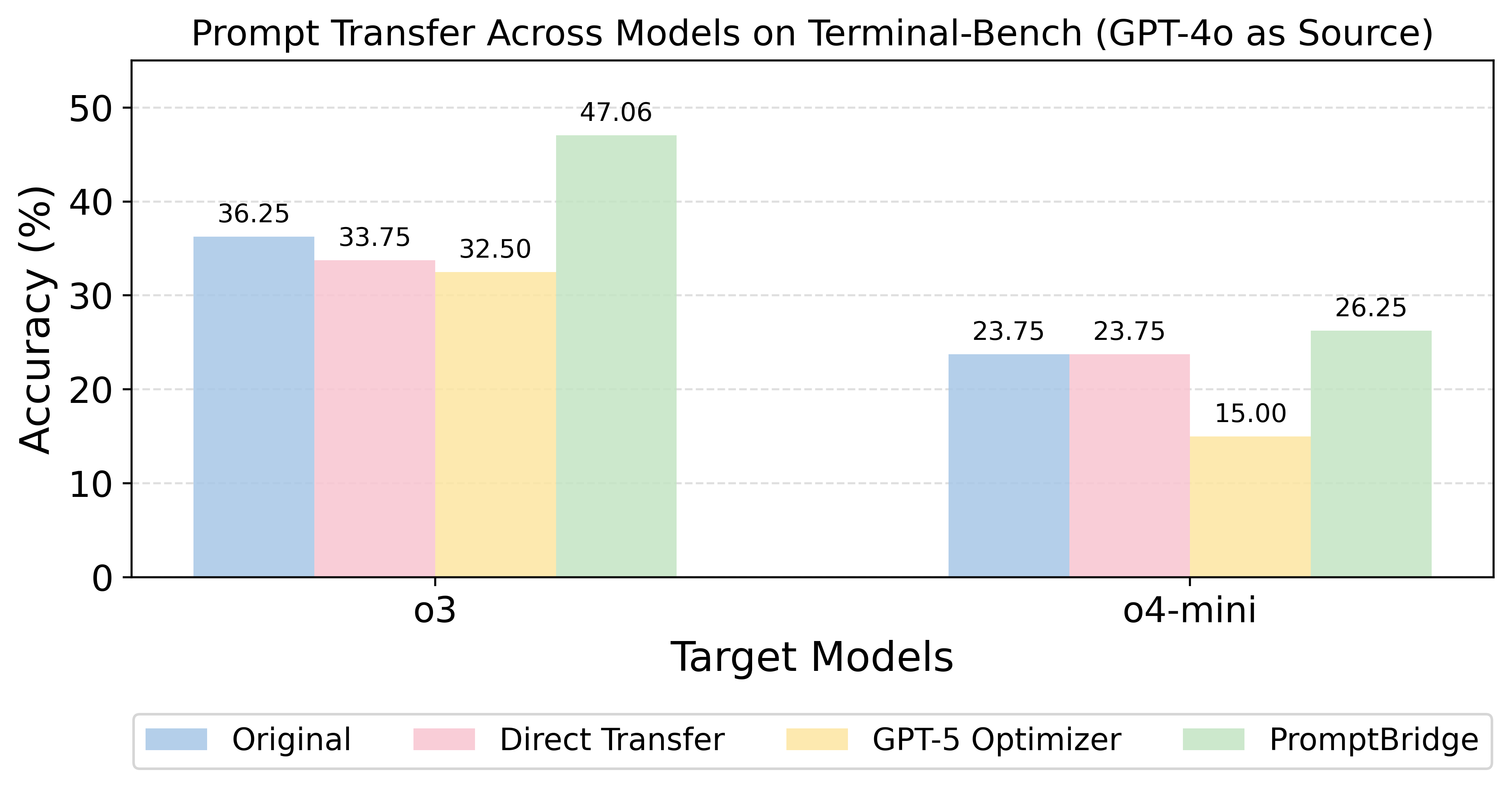}
    % \vspace{-15pt}
    \caption{\textbf{Results on Terminal-Bench.} We use \texttt{GPT-4o} as the source model and evaluate transferability on \texttt{o3} and \texttt{o4-mini} as target models. On the source, the accuracy with the default prompt is 15\%, which increases to 18.75\% after calibration with the optimized prompt.
    Original denotes the baseline performance using the default prompt template from Terminus. Direct Transfer refers to directly applying the optimized prompt from \texttt{GPT-4o} to the other models without further adaptation. }
    \label{fig:terminal-results-gpt4o}
\end{figure}

\subsection{Prompt Transfer}
\label{exp:prompt_transfer}

\paragraph{Single-Agent Setting. } 
\autoref{tab:gpt4o-main-code} presents the generalization performance of the proposed \method across unseen coding tasks and target LLMs, demonstrating its effectiveness in transferring optimized prompts between models. The first row reports results on the test datasets using the source model, \texttt{GPT-4o}. The target models, \texttt{o3}, \texttt{o4-mini}, and \texttt{Llama-3.1-70B-Instruct}, represent three practical deployment scenarios: a more capable model, a cost-efficient model, and an open-source alternative, respectively. 
For relatively simple code generation tasks (HumanEval, MBPP), \method achieves the highest accuracy.
On \texttt{o3}, GEPA performs comparably on MBPP but does not consistently match \method’s effectiveness. 
Across more challenging problem-solving benchmarks such as APPS, xCodeEval, and CodeContests, \method demonstrates substantial improvements over direct transfer, achieving relative gains of 11.5\%, 13.33\%, and 15.92\% for \texttt{o3}; 0.58\%, 7.86\%, and 46.54\% for \texttt{o4-mini}; and 11.1\%, 14.73\%, and 18.02\% for \texttt{Llama-3.1-70B-Instruct}, respectively.
Since the baseline methods (GPT-5 Optimizer and MIPROv2) are not model-adaptive, their performance remains inconsistent, occasionally competitive but lacking robustness across tasks. GEPA, which requires a task-specific LLM, can implicitly encode model preferences. However, as GEPA directly applies the optimized prompt learned from the training dataset to the test benchmark, it risks overfitting to training-specific features, resulting in reduced generalization. 
Overall, the results across diverse models and benchmarks highlight the robustness and generalization capability of the proposed transfer method.
More results can be found in~\autoref{tab:app-gpt4o-others} in the Appendix.

\autoref{fig:swe-results} presents the transfer results on SWE-Bench Verified. \method consistently improves performance over direct transfer, demonstrating its effectiveness in adapting optimized prompts across models. The optimized prompt for \texttt{o4-mini} achieves a 9\% gain compared with the default prompt. 
\method achieves substantial improvements over direct transfer, with relative gains of 27.39\% for \texttt{o3}, 15.79\% for \texttt{Llama-3.1-70B-Instruct}, and 2.5\% for \texttt{GPT-4o}. However, the GPT-5 Optimizer, despite being a strong baseline, fails to deliver stable improvements. On \texttt{GPT-4o}, its performance even falls below direct transfer, with only 9.40\%.
\autoref{fig:terminal-results-gpt4o} presents the transfer results on Terminal-Bench. 
When transferred to different target models, \method achieves relative improvements of 39.44\% on \texttt{o3} and 10.53\% on \texttt{o4-mini} compared with the direct transfer baseline. 
In contrast, the GPT-5 Optimizer performs poorly when handling cross-model transfer without leveraging transferable knowledge, even yielding worse performance than direct transfer. 
On \texttt{o3}, the performance of direct transfer does not improve over original prompt. This finding suggests that a prompt optimized for one model or system may not remain optimal across different LLMs, underscoring the importance of discovering model-adaptive prompts for robust cross-model performance. These results further demonstrate the effectiveness and necessity of our framework in mitigating performance degradation across models. More analysis on SWE-Bench and Terminal-Bench can be found in~\autoref{app:results:swe} and \autoref{app:results:terminal} in the Appendix, respectively.

\begin{table*}
\centering
\caption{Pass@1 Accuracy on HumanEval using different calibration methods in multi-agent setting (MapCoder) when model drifting occurs local or global. Source model is \texttt{GPT-4o}, target model is \texttt{Llama3.1-70B-Instruct}. Global Agents denotes that all agents in this multi-agent system switch from source model to target model; Local Coding Agent denotes that the coding agent switches from source model to target model.} 
\resizebox{\textwidth}{!}{
\begin{tabular}{c|c|ccc}
\toprule
 \multicolumn{1}{c}{\textbf{Method}}  &\multicolumn{1}{c}{\textbf{Global Model Drifting}} & \multicolumn{3}{c}{\textbf{Local Model Drifting}}  \\ 
\midrule
\textbf{Dataset} & All agents  & \multicolumn{1}{c}{Coding Agent}  & \multicolumn{1}{c}{Planning Agent}  & Debugging Agent  \\ 
\midrule
Source Model  &92.68 & 92.68 & 92.68 & 92.68 \\
\midrule
\multicolumn{5}{c}{\textbf{Target Model: Llama3.1-70B-Instruct}} \\
\midrule
Direct Transfer  &  87.59 & \textbf{90.24} & \textbf{92.07}& 88.41\\
GPT-5 Optimizer & \textbf{87.80}  &  87.80  & 90.24 & 85.96 \\
\method (Ours) & \textbf{87.80} & 89.63 & 87.20 & \textbf{89.02} \\
\midrule
\multicolumn{5}{c}{\textbf{Target Model: o3}} \\
\midrule
Direct Transfer& 94.51  & 94.51& 95.73& 93.29\\
GPT-5 Optimizer &93.29 & \textbf{94.51} & \textbf{96.34} & 94.51\\
\method (Ours) & \textbf{95.73} & 93.90 & 95.12 & \textbf{95.12}  \\
\bottomrule
\end{tabular}
}
\label{tab:mapcoder-results}
\end{table*}

\begin{table*}
\centering
\caption{Results on TravelPlanner validation set under Two Stage Mode. It involves two agent: React Agent and Planner Agent. Global Model Drifting denotes that all agents switch from the original model to the target model; Local Model Drifting denotes that the planner agent switches from the source model to the target model.} 
\resizebox{\textwidth}{!}{
\begin{tabular}{c|c|cc|cc|c}
\toprule
\textbf{Metric} & \multicolumn{1}{c}{\textbf{Final Pass Rate}}& \multicolumn{2}{c}{\textbf{CC Pass Rate}}& \multicolumn{2}{c}{\textbf{HC Pass Rate}}  & \multicolumn{1}{c}{\textbf{Delivery Rate}}   \\ 
\midrule
\multicolumn{1}{c}{\textbf{Method}} &   &\multicolumn{1}{c}{\textbf{Micro}}    & \multicolumn{1}{c|}{\textbf{Macro}} &\multicolumn{1}{c}{\textbf{Micro}}    & \multicolumn{1}{c|}{\textbf{Macro}} & \\ 
\midrule
Source Model: GPT-4o (Original) & 0.0 & 50.01 & 2.22 & 7.62 & 2.78 & 66.67 \\
Target Model: o3 (Original) & 1.67 & 41.04 & 2.22 & 12.14 & 10.56 & 58.89\\
\midrule
\multicolumn{7}{c}{\textbf{Global Model Drifting}} \\
\midrule
Source Model: GPT-4o (Optimized) & 2.78 & 61.25 & 5.56 & 19.52 & 15.56 & 83.33\\
\midrule
Frozen Direct Transfer   & 0.56 & \textbf{63.96} & 1.11 & \textbf{28.81} & \textbf{26.67} & 82.78\\
\method - optimized & \textbf{2.22} & 61.46 & \textbf{4.44} & 18.81 & 15.56 & \textbf{85.56} \\
\midrule
\multicolumn{7}{c}{\textbf{Local Model Drifting (Planner)}} \\
\midrule
Source Model: GPT-4o (Optimized) & 3.33 & 60.56 & 4.44 & 11.67 & 13.33 & 85.56\\
\midrule
Frozen Direct Transfer & 1.11 & 67.64 & 3.89 & 28.81 & 23.33 & \textbf{87.78} \\
\method - optimized & \textbf{2.78} & \textbf{67.85} & \textbf{4.44} & \textbf{38.33} & \textbf{30.00} & 86.67 \\
\bottomrule
\end{tabular}
}
\label{tab:travelplanner-agent}
\end{table*}

\begin{table*}
\centering
\caption{Pass@1 accuracy on multiple code generation datasets using different calibration methods, evaluated on \texttt{o3} and \texttt{o4-mini}. Frozen Prompt denotes directly applying the prompt of \texttt{GPT-4o} to the models without adaptation. It can also be viewed as a default prompt. Average denotes the mean accuracy across the five code benchmarks, and higher values indicate better performance.
Higher values indicate better performance. The best results are highlighted in \textbf{bold}.  } 
\resizebox{0.9\textwidth}{!}{
\begin{tabular}{c|ccccc|c}
\toprule
 \multicolumn{1}{c}{\textbf{Method}} & \multicolumn{1}{c}{\textbf{HumanEval}}  & \multicolumn{1}{c}{\textbf{MBPP}} & \multicolumn{1}{c}{\textbf{APPS}}    & \multicolumn{1}{c}{\textbf{xCodeEval}} & \multicolumn{1}{c|}{\textbf{CodeContests}} &   \multicolumn{1}{c}{\textbf{Average}} \\ 
\midrule
\multicolumn{7}{c}{\textbf{Model: o3}} \\
\midrule
Frozen Prompt & 92.27 & 77.92 & 32.67  & 66.04 & 48.61 & 63.50\\
MIPROv2* & 97.56 & 47.52 & 36.00 & 71.38 & 40.00 & 58.49 \\
GEPA* & 98.17 & \textbf{83.79} & 33.11& 69.18 & 46.87 & 66.22\\
MAP-RPE& \textbf{98.37} & 80.86 & \textbf{36.67} & \textbf{74.53} & \textbf{58.79} & \textbf{69.84} \\
\midrule
\multicolumn{7}{c}{\textbf{Model: o4-mini}} \\
\midrule
Frozen Prompt & 96.54 & 79.09 & 37.78 & 72.01 & 40.12 & 65.11 \\
MIPROv2* & 98.78 & 15.95 & 38.66 & \textbf{78.30} & 32.73 & 52.88 \\
GEPA* & \textbf{98.78} & 79.26 & 34.89 & 76.73 & \textbf{56.23} & \textbf{69.18} \\
MAP-RPE & 96.95 & \textbf{79.60} & \textbf{39.33} & 76.42 & 47.88 & 68.04  \\
\bottomrule
\end{tabular}
}
\label{tab:optimizer}
\end{table*}

\paragraph{Multi-Agent System. } \autoref{tab:mapcoder-results} reports the Pass@1 accuracy on the HumanEval dataset in a multi-agent setting under both local and global model drifting scenarios.
For global model drifting, \method consistently improves performance compared with direct transfer, whereas the GPT-5 Optimizer does not yield consistent gains across different target LLMs.
For local model drifting, where only specific agents (e.g., coding or planning agents) are switched, \method is not always the top performer; however, for the debugging agent, it consistently achieves the best results. This observation aligns with the observation that the debugging agent plays a central role in the overall multi-agent workflow as indicated in \citet{islam2024mapcoder}.
In the case of global model drifting, \method jointly updates all four agent prompts, which may enhance coherence among agents and lead to superior performance.
Although the overall performance improvement is not large, our emphasis is on maintaining performance stability across different LLMs after transfer, initially demonstrating that \method mitigates degradation when model drift occurs in the multi-agent system.

\paragraph{Non-Coding Domain.}  
We also conduct experiments in the non-coding domain.  \autoref{tab:travelplanner-agent} presents the results under the two-stage mode. The source model is \texttt{GPT-4o}, and the target model is \texttt{o3}. Using the default prompt leads to pass rates of 0.0\% and 1.67\%, respectively. When the optimized prompts are applied to both the React and Planner agents, the pass rates increase: Global reaches 2.78\%, and Local achieves 3.33\%. In this case, optimizing the React prompt slightly decreases performance.
\method achieves consistent improvements over direct transfer and outperforms the default prompt configuration. Notably, even in this non-coding setting, \method enhances performance despite being trained exclusively on coding-domain datasets.
The results of TravelPlanner under the sole-planning mode are provided in the~\autoref{app:results:travelplanner} in the Appendix.

\subsection{Prompt Calibration}
\label{exp:prompt_calibration}
% \paragraph{Results on Prompt Calibration.}
Prompt Calibration aims to generate model-adaptive and task-specific optimal prompts. To evaluate the effectiveness of different optimization strategies, we compare our proposed MAP-RPE with several existing methods, including GEPA and MIPROv2, using five code generation benchmarks. As shown in \autoref{tab:optimizer}, we evaluate optimized prompts on \texttt{o3} and \texttt{o4-mini} as target specific models.
For the \texttt{o3}, MAP-RPE consistently yields high accuracy across all datasets and has the highest average accuracy, achieving the best performance on xCodeEval (74.53) and CodeContests (58.79), indicating that our method effectively tailors prompts to model-specific behavior. On \texttt{o4-mini}, MAP-RPE maintains balanced performance, achieving the top results on MBPP and APPS, while avoiding the instability seen in MIPROv2. It achieves average accuracy comparable to that of GEPA.
Across all settings, MAP-RPE consistently delivers strong and stable performance, outperforming frozen prompt and achieving competitive or superior results compared with prior SOTA prompt optimization frameworks. These results demonstrate that reflective, metric-guided evolution enables robust calibration of prompts to model-specific and task-specific characteristics, forming the foundation for reliable prompt adaptation within the proposed PromptBridge framework.

\subsection{Ablation Study}

\begin{table*}
\centering
\caption{Pass@1 Accuracy on several code generation datasets when switching the original model \texttt{GPT-4o} to the target models, including \texttt{o3}, \texttt{o4-mini}, \texttt{Llama3.1-70B-Instruct}. Direct Transfer refers to apply the source prompt to the target model directly. The higher the accuracy the better. The best transfer results are highlighted in \textbf{Bold}.
} 
\resizebox{0.85\textwidth}{!}{
\begin{tabular}{c|ccccc}
\toprule
 \multicolumn{1}{c}{\textbf{Method}} & \multicolumn{1}{c}{\textbf{HumanEval}}  & \multicolumn{1}{c}{\textbf{MBPP}} & \multicolumn{1}{c}{\textbf{APPS}}    & \multicolumn{1}{c}{\textbf{xCodeEval}} & \multicolumn{1}{c}{\textbf{CodeContests}}\\ 
\midrule
Source Model: GPT-4o & 91.10 & 79.80 & 12.00 & 37.03 & 5.45 \\
\midrule
\multicolumn{6}{c}{\textbf{Target Model: o3}} \\
\midrule
Direct Transfer & 92.27 & 77.92 & 32.67  & 66.04 & 48.61\\
One-shot ICL & 89.63 &79.43& 33.11 & 71.66  & 52.32\\
Few-shot ICL  & 87.40 & 77.70 & \textbf{36.89} & 71.38 & 51.63\\
\rowcolor{gray!30}  
\method (Ours) &\textbf{97.15} & \textbf{80.44} & 36.44 & \textbf{74.84} & \textbf{56.36} \\
\midrule
\multicolumn{6}{c}{\textbf{Target Model: o4-mini}} \\
\midrule
Direct Transfer & 96.54 & 79.09 & 37.78 & 72.01 & 40.12 \\
One-shot ICL & 97.56 & 79.18 & 35.33 & 76.10 & 57.78\\
Few-shot ICL & 97.56 & 80.52 & 37.33  & 77.36 & 57.17 \\
\rowcolor{gray!30}  \method (Ours) & \textbf{98.37} & \textbf{80.60} & \textbf{38.00} & 77.67 & \textbf{58.79} \\
\midrule
\multicolumn{6}{c}{\textbf{Target Model: Llama3.1-70B-Instruct}} \\
\midrule
Direct Transfer & 68.70 & 65.57 & 10.00 & 21.38 & 20.20\\
One-shot ICL & 41.26& 36.78 & 10.22 & 19.18 & 22.83 \\
Few-shot ICL & 78.46 & 72.04 & 5.11 & 21.70 & 3.64 \\
\rowcolor{gray!30}  \method (Ours) & \textbf{79.88} & \textbf{73.64} & \textbf{11.11} & \textbf{24.53} & \textbf{23.84}\\
\bottomrule
\end{tabular}
}
\label{tab:ablation-transfer}
\end{table*}

\paragraph{Ablation Study: Understanding Cross-Model Prompt Transferability }
To further evaluate the effectiveness of PromptBridge, we conduct an ablation study comparing three transfer strategies: 1) \emph{One-shot in-context learning (ICL)}, which uses the single highest-scoring example source-target pair as the demonstration; 2) \emph{Few-shot ICL}, which uses the top five pairs as demonstrations; and 3) our \emph{PromptBridge}, which first distills the systematic prompt differences across alignment tasks and then applies this mapping to unseen tasks.
\autoref{tab:ablation-transfer} shows that one-shot ICL and few-shot ICL provide only limited improvements and are often unstable due to their high sensitivity to the specific example pairs used during in-context conditioning. This instability likely arises because the model imitates superficial, task-specific patterns in the provided examples rather than learning a generalizable transformation between models. In contrast, PromptBridge consistently achieves the highest transfer performance. By summarizing cross-model prompt differences into an explicit transformation rule, it captures stable and reusable adaptation patterns that generalize beyond the observed examples. These results demonstrate that learning a high-level, model-level transformation is substantially more robust and effective than relying solely on direct ICL examples.

\begin{figure}[t]
    \centering
    \includegraphics[width=0.9\textwidth]{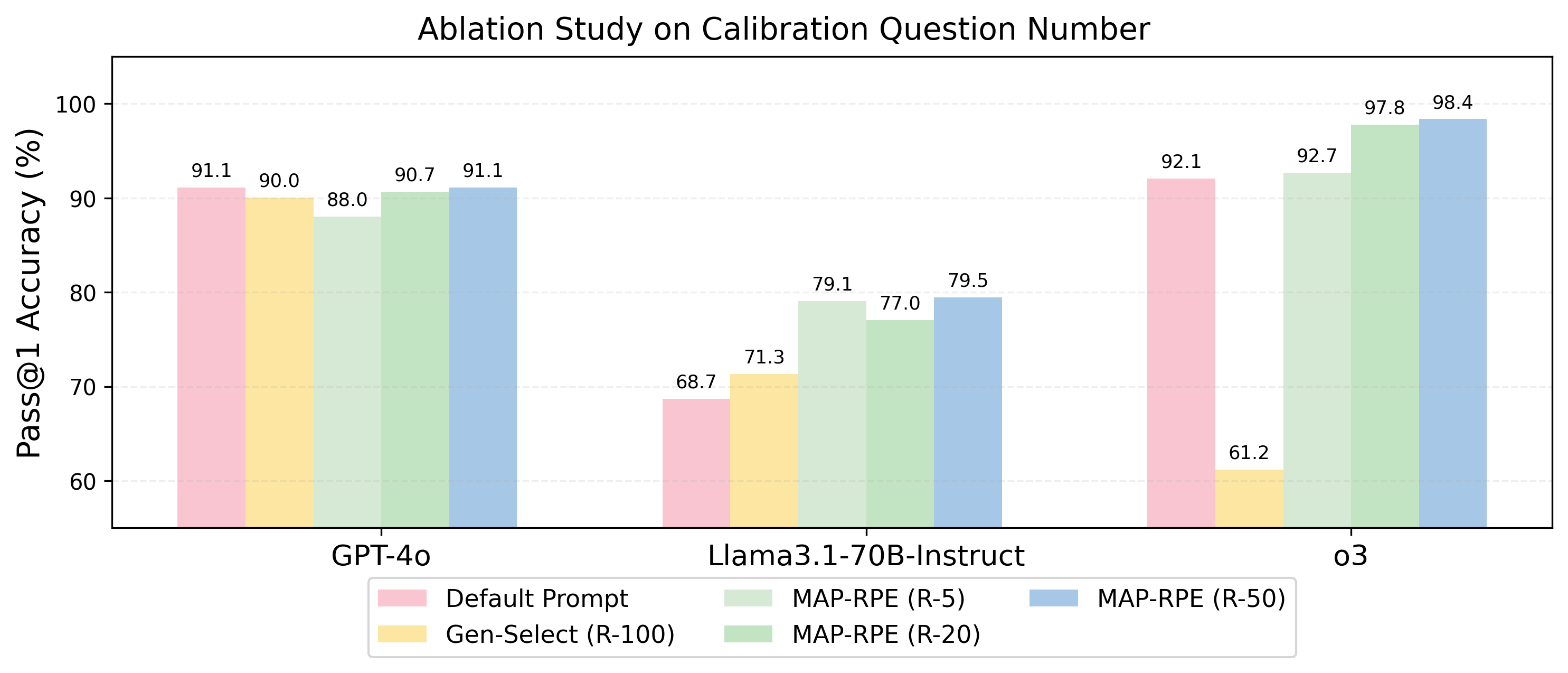}
    % \vspace{-15pt}
    \caption{\textbf{Ablation study on the number of calibration questions.} Pass@1 accuracy on HumanEval using different calibration methods to derive the optimal prompt for each model. Default Prompt denotes the original designed prompt for solving coding tasks. Gen-Select refers to generating multiple prompt candidates and selecting the best-performing one.}
    \label{fig:ablation-question}
\end{figure}

\paragraph{Ablation study on the calibration question number.}
\autoref{fig:ablation-question} reports the Pass@1 accuracy on HumanEval under different calibration strategies (Generation–Selection, see~\autoref{app:setting:baseline} in the Appendix) and varying numbers of calibration questions. Across all models, MAP-RPE consistently achieves the highest accuracy, demonstrating its effectiveness in adaptively aligning prompt semantics to model-specific behaviors. For instance, on \texttt{Llama3.1-70B-Instruct}, MAP-RPE achieves up to 79.5\% Pass@1, significantly outperforming the default prompt baseline (68.7\%); on \texttt{o3}, MAP-RPE reaches 98.4\%.
The ablation on the number of calibration questions ($n \in {5, 20, 50}$) further illustrates that performance improves steadily as $n$ increases, since a larger calibration set exposes the optimizer to a more diverse range of reasoning and formatting patterns. Notably, MAP-RPE is sample-efficient: even with only five calibration questions, it already achieves substantial gains over the default prompt on \texttt{Llama3.1-70B-Instruct} and \texttt{o3}. 
Compared with random Generation–Selection, which often exhibits unstable performance and may occasionally find a locally optimal candidate, MAP-RPE remains consistently strong, highlighting the advantage of reflective, metric-guided calibration and systematic adaptivity over heuristic search.

\section{Conclusions}
We investigated the overlooked challenge of \emph{prompt transfer under model drifting}: the degradation that occurs when prompts tuned for one LLM are reused on another. We formalized this problem by empirical analysis and introduced a practical framework, PromptBridge, that (1) evolves task- and model-specific prompts and (2) learns cross-model transfer mapping to translate prompts for unseen tasks when switching models. Across single-agent and multi-agent settings, our experiments show consistent gains in performance, indicating that efficient prompt transfer is important for sustainable LLM development. 
Treating prompts as migratable artifacts rather than static instructions offers a practical path to maintain performance as LLM ecosystems evolve. Our framework provides a new paradigm to make model switching more reliable and efficient.

\noindent\textbf{Future Work.} In future, we plan to invest in research across the following directions.
\begin{itemize}
    \item \textbf{Broader Model Families and Alignment-aware Variants.} Extending the study to additional model families, such as Mistral and DeepSeek, and to alignment-aware variants (e.g., SFT-, DPO-, or GRPO-aligned models) would provide a more complete characterization of how alignment methods reshape prompt patterns. 
    \item \textbf{Behavioral and Stylistic Drift.}  Beyond performance, large models often exhibit noticeable shifts in style, verbosity, structure, or safety behavior when moving across models (e.g., \texttt{GPT-4o} vs.~\texttt{GPT-5}). We plan to conduct
    a systematic study of how stylistic and behavioral patterns drift, and how such drift affects prompt robustness.
    \item \textbf{Community Benchmarks for Prompt Portability.}  
    We plan to develop shared benchmarks and standardized corpora specifically designed to evaluate cross-model prompt portability across families, scales, and alignment settings. 
\end{itemize}

\newpage

\raggedright
\bibliography{main}
\bibliographystyle{main}

\newpage
\appendix
\justifying

\startcontents[appendix]
\printcontents[appendix]{l}{1}{\section*{Appendix Contents}}

\section{Preliminary Details}

\subsection{Necessity of Cross-Model Prompt Transfer}
\label{app:pre:necessity}
\paragraph{Detailed setup for \autoref{fig:pre}.} For each model, we first apply our prompt evolution method to obtain the optimal prompt template, using a calibration number of 50. In other words, the results of MAP-RPE (Random-50) in \autoref{tab:optimizer} correspond to the diagonal entries in figure (b). The reported results represent the average accuracy over three runs. When transferring the optimized prompt from a source model to a target model, the transferred prompt does not necessarily yield the best performance. We also conduct experiments within the same GPT family.

\begin{figure}[t]
    \centering
    \includegraphics[width=1.0\textwidth]{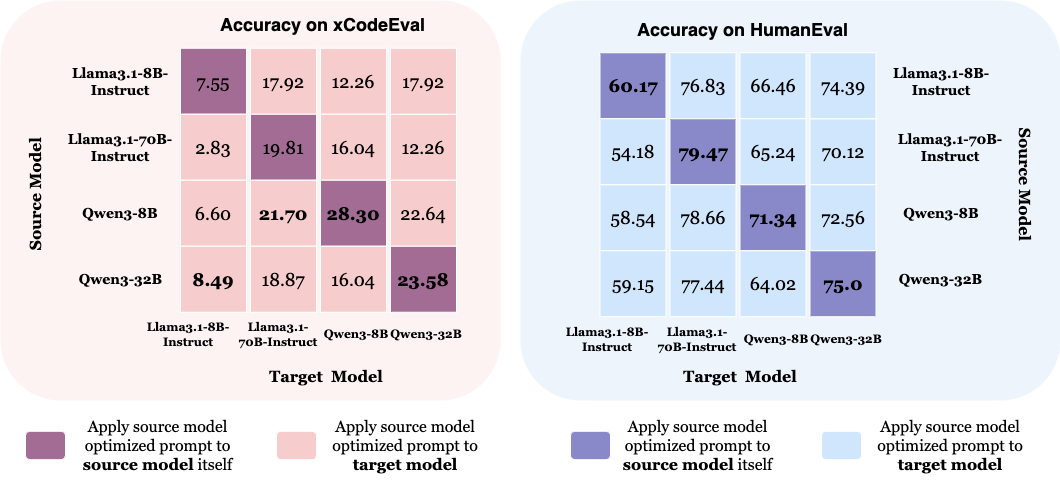}
    \vspace{-0.2cm}
    \caption{\textbf{Cross-model frozen prompt transfer on xCodeEval (left) and HumanEval (right).} Each row shows the source model and each column the target model. Darker diagonal cells indicate the accuracy obtained when applying a model’s own optimized prompt, while lighter off-diagonal cells show performance when transferring that optimized prompt to a different model without further adaptation. The consistent diagonal–off-diagonal gap reveals substantial model drifting: prompts carefully optimized for one model fail to generalize to others, even within the same family. 
    }\label{fig:qwen-llama}
    \vspace{-0.4cm}
\end{figure}

\paragraph{Model Drifting Demonstration} 
Beyond the results shown in \autoref{fig:pre}, we further evaluate cross-model transfer within the same model family but at different scales. As shown in \autoref{fig:qwen-llama}, both xCodeEval and HumanEval exhibit a consistent pattern of model drifting, even when source and target models share the same architecture. In nearly all cases, the diagonal entries, where a model uses its own optimized prompt, achieve the highest accuracy, while transferring that prompt to other scales within the same family leads to substantial degradation (e.g., \texttt{Llama3.1-70B-Instruct} → \texttt{Llama3.1-8B-Instruct}, \texttt{Qwen3-32B} → \texttt{Qwen3-8B}). The drop can be dramatic, reaching up to 50–70\% relative loss on xCodeEval, indicating that the optimal prompt for a model does not smoothly scale across model sizes. Although HumanEval displays higher absolute accuracy, the same diagonal–off-diagonal gap persists, confirming that even closely related models do not share a stable prompt optimum. These results demonstrate that model drifting arises not only across different model families, but also within a single family across scales, further underscoring the need for model-adaptive prompt transfer mechanisms such as \method.

\textbf{Default prompt in Multi-Agent System } 
Model drifting is not limited to optimized prompts tailored for specific source models; it can also occur with default, human-designed prompts that are intended to be universal across models but are not guaranteed to be optimal for any of them. 
Here, we evaluate the default performance using the original prompt templates for \texttt{GPT-4o}, \texttt{Llama3.1-70B-Instruct}, and other models. Then we use GPT-5 optimizer to refine the debugging agent’s prompt, producing an optimized version for comparison. 
Although default prompt templates are typically designed by humans to work broadly across models, they are seldom optimal for each specific model (see \autoref{tab:mapcoder-results}). Since model drifting is defined as the performance change caused by substituting the underlying model, such universal designs without optimization can still exhibit model drifting when directly transferred. This observation underscores the importance of applying targeted prompt transfer or adaptation to existing prompt templates when deploying systems across different LLMs.

\subsection{Feasibility of Cross-Model Prompt Transfer}
\label{app:pre:feasibility}
We hypothesize that there exists a shared pattern of prompt change that generalizes across tasks when transferring between models.
To examine this, we study whether the edit patterns required to transform an optimal source prompt into its target counterpart remain consistent across tasks.
If such pattern exists, it would indicate that model adaptation follows a stable semantic trajectory, suggesting that prompts can be transferred between models through shared transformation principles rather than being re-optimized from scratch for every task. 
We evaluate this feasibility from semantic perspectives, analyzing how prompt deltas behave across models and tasks.

\begin{figure}[t]
    \centering
    \begin{subfigure}[t]{0.48\textwidth}
        \centering
        \includegraphics[width=\textwidth]{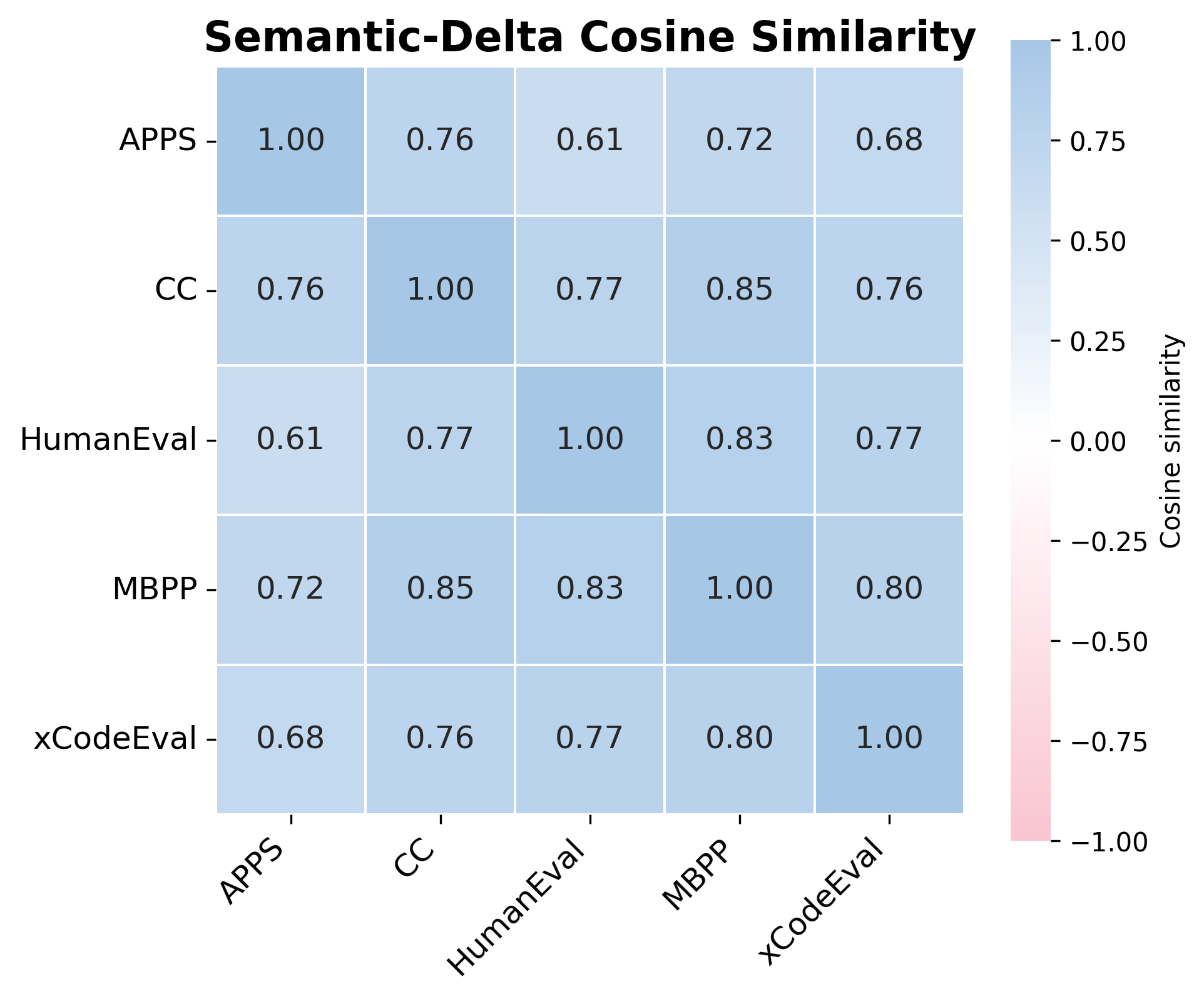}
        \caption{GPT4o→Llama-3.1-70B-Instruct.}
        \label{fig:preliminary-cosine-llama3}
    \end{subfigure}
    \hfill
    \begin{subfigure}[t]{0.48\textwidth}
        \centering
        \includegraphics[width=\textwidth]{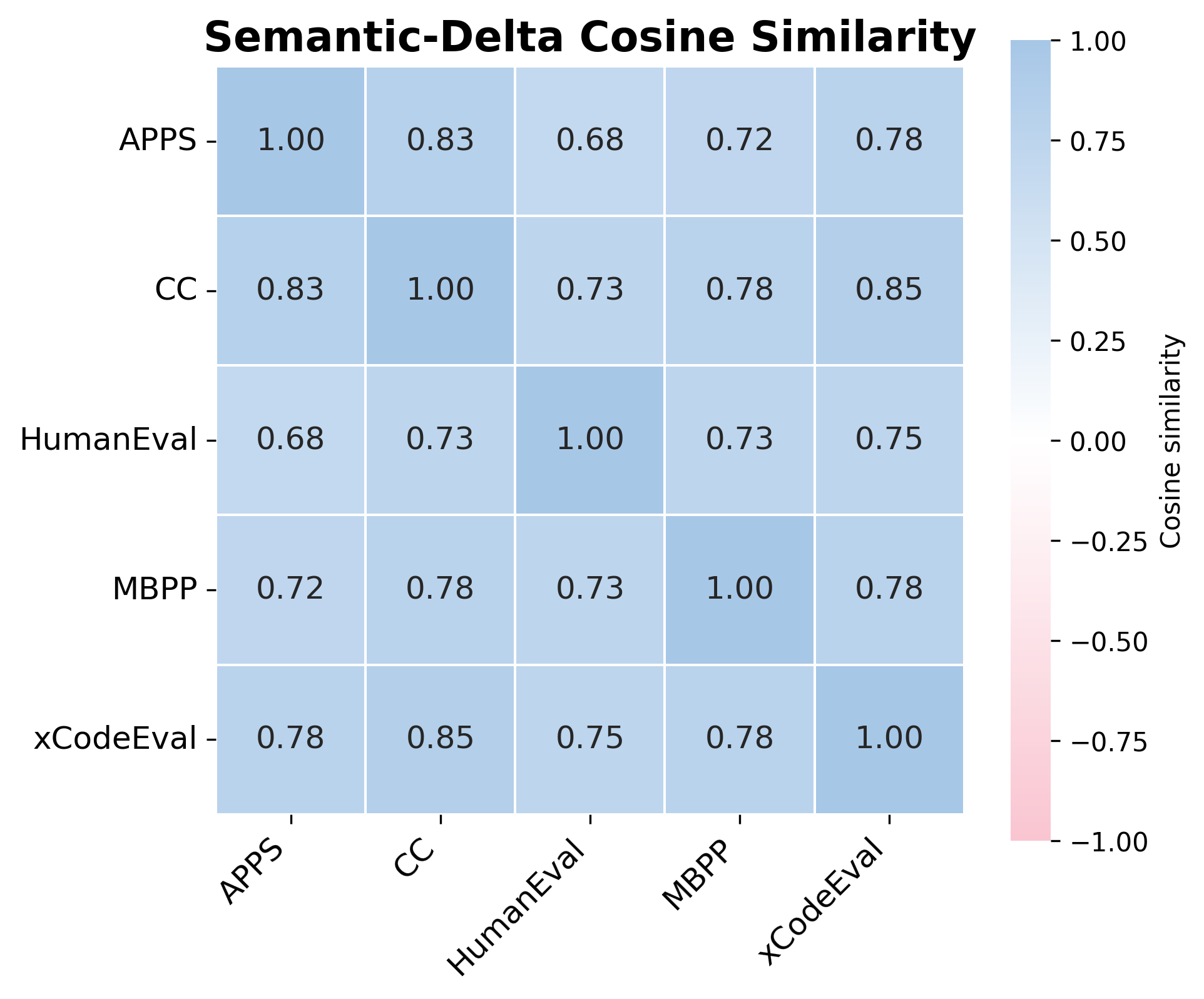}
        \caption{GPT4o→o3.}
        \label{fig:preliminary-cosine-o3}
    \end{subfigure}

    \caption{\textbf{Semantic-delta cosine similarity across tasks for cross-model prompt transfer.} We compute the semantic-delta embeddings which captures how each task’s output distribution shifts under prompt transfer, and measure the cosine similarity between tasks. Higher similarity indicates shared transfer patterns. (a) compares \texttt{GPT-4o} → \texttt{Llama-3.1-70B-Instruct}; (b) compares \texttt{GPT-4o} → \texttt{o3}.}
    \label{fig:preliminary-semantic}
\end{figure}

\textbf{Experimental Setup } We consider fixed source→target pairs (e.g., from \texttt{GPT-4o} to \texttt{o3}; from \texttt{GPT-4o} to \texttt{Llama-3.1-70B-Instruct}). For each pair we hold both models fixed during analysis. We use five code generation benchmarks, which are unseen tasks in the main text. Here, our goal is not to evaluate task performance itself but to analyze the pattern changes; thus, the specific tasks are not the focus.
For each task $t$, we construct $\mathcal{P}^*_{\textrm{source},t}$, which refers to the source-optimal prompt found on the source model; $\mathcal{P}^*_{\textrm{target},t}$, which refers to the target-optimal prompt found on the target model. We compute the Semantic delta $v_t=e(\mathcal{P}^*_{\textrm{target},t})-e(\mathcal{P}^*_{\textrm{source},t})$, where $e(\cdot )$ is a fixed text embedding (here we use all-MiniLM-L6-v2 model\footnote{https://huggingface.co/sentence-transformers/all-MiniLM-L6-v2}) We compute pairwise cosine similarities 
$cos(v_i,v_j)$ for all task pairs $i \neq j$, higher values indicate a shared direction of edits from source to target.

\textbf{Semantic Delta Embedding Similarity } 
\autoref{fig:preliminary-semantic} presents the pairwise cosine similarity matrices of semantic deltas across five tasks for two source–target model pairs. Blue regions indicate higher alignment between task-specific prompt adaptations. The consistently positive similarities reveal that source-to-target prompt transformations exhibit shared latent structures, suggesting the existence of a transferable pattern across tasks and thus supporting the feasibility of cross-task prompt transfer.
However, embedding similarity alone does not imply functional equivalence. To further validate feasibility, we assess performance changes when applying \method to unseen tasks, demonstrating that learned transfer effects can generalize beyond the observed task space.
While our current work primarily focuses on establishing the model drifting problem and developing a general cross-model prompt transfer framework, these findings hint that incorporating task-level adaptation could further enhance transfer effectiveness. Exploring such cross-task prompt transfer represents an interesting direction for future work to make prompt transfer more powerful and context-aware.

\section{Experimental Settings}
\label{app:setting}
\subsection{Baseline Methods}
\label{app:setting:baseline}
In this section, we introduce all the baselines used in the experiments. 

\paragraph{GPT5 Optimizer} For the five coding benchmarks, we provide GPT-5 with the instruction shown in \autoref{lst:gpt5-optimizer}, prompting it to generate an optimized prompt based on the original one. Note that this process is essentially zero-shot, as the model is only given the name of the target model without any additional examples or training data.

\begin{lstlisting}[caption={The prompt used in GPT5 Optimizer baseline for five coding datasets.}, label=lst:gpt5-optimizer]
You are an expert in optimizing prompts for code generation tasks - {dataset}. Your task is to optimize the following prompt for generating Python code that solves a specific problem. The original prompt is provided below.
Your goal is to make the prompt more concise, clear, and effective for the {specific_model} model, while ensuring it remains understandable and retains the original intent. You may add additional guidance or context that you believe will help the model generate better responses.
## ====== Original Prompt Starts ========
## Original Prompt: {original_prompt}
## ====== Original Prompt Ends ========
Please provide an optimized version of the prompt that is suitable for the {specific_model} model and can elict better response.

Optimized Prompt:
\end{lstlisting}

\paragraph{MIPROv2~\citep{opsahl2024optimizing}} is a widely used prompt optimizer and has been integrated into the DSPy~\citep{khattab2024dspy} framework. It works by jointly optimizing both instructions and demonstrations using Bayesian optimization. For each problem module, it first bootstraps candidate sets of instructions and demonstrations, assigning uniform priors over their utilities. Candidate assignments are proposed with the Tree-Structured Parzen Estimator (TPE), and the Bayesian model is updated based on evaluation scores to favor high-performing candidates. The most probable sets of instructions and demonstrations are then selected and validated to obtain the final optimized program configuration. 
All MIPROv2 optimization runs are performer with the $auto=heavy$ setting. For our training dataset, we use a textual similarity score between the prediction and the ground truth response, serving as a simple evaluation metric. We combine all the training dataset into a whole one to perform the optimizer. 

\paragraph{GEPA~\citep{agrawal2025gepa}} is a prompt optimizer that thoroughly incorporates natural language reflection to learn high-level rules from trial and error. It works iteratively-proposing a new candidate in every iteration by improving some existing candidates using either reflective prompt mutation or system aware merge. To avoid local optimum, it introduces Pareto-based candidate sampling, which filters and samples from the list of best candidates per task. 

\paragraph{MIPROv2* and GEPA*} denote the variants where the optimizers are applied on the test benchmark. Specifically, we randomly select 50 questions from the test dataset and use them to derive the optimized user instruction. All other parameters are kept identical to those used in MIPROv2 and GEPA, respectively.

\paragraph{Calibration: Generation-Selection~\citet{wang2023promptagent}}
Given $n$ questions from a task $T_i$, generate the optimal prompt $p_i$ for the task on target model. Generation-selection is a navie method, first we utilize GPT-5 to generate several candidate prompts, then we use the similarity between the generated response and the original response which is generated by the source model as a selection metric to select those candidates with highest similarity score.

\subsection{Dataset Description}
\label{app: dataset_description}

\begin{table*}
\centering
\caption{Alignment Tasks and Unseen Tasks information.} 
\resizebox{0.7\textwidth}{!}{
\begin{tabular}{ccc}
\toprule
 \multicolumn{1}{c}{\textbf{Dataset Name}} & \multicolumn{1}{c}{\textbf{Domain}} & \multicolumn{1}{c}{\textbf{Data Number}}  \\ 
\midrule
\multicolumn{3}{c}{Alignment Tasks} \\
\midrule
synthetic-code-generations~\citep{wei2023magicoder}& Coding & - \\
CodeContests (Train)~\citep{li2022competition} & Coding  & -  \\
\midrule
\multicolumn{3}{c}{Unseen Tasks} \\
\midrule
HumanEval~\citep{chen2021evaluating} & Coding & 164\\
MBPP~\citep{austin2021program} & Coding & 397  \\
APPS~\citep{hendrycks2021measuring} & Coding & 150 \\
xCodeEval~\citep{khan2023xcodeeval} & Coding & 106 \\
CodeContests~\citep{li2022competition} & Coding & 156 \\
SWE-bench Verified~\cite{jimenez2024swebench} & Agent & 500  \\
Terminal Bench~\citep{tbench_2025} & Agent & 80 \\
TravelPlanner - Validation~\citep{xie2024travelplanner} & Planning & 180 \\
% GAIA & \\
\bottomrule
\end{tabular}
}
\end{table*}

% deepmind/code_contests
For the training dataset, we randomly sample 30 subtasks from the synthetic-code-generation dataset, with each subtask containing 100 question–answer pairs. For the CodeContests dataset, we select 24 tasks based on their difficulty levels.
For each subtask or task, we apply the proposed MAP-RPE method to calibrate the optimal prompt and record its corresponding evaluation results. In total, this yields 30 + 24 = 54 alignment tasks used for calibration.
To evaluate the performance, we use five coding benchmark datasets: two from basic programming and three from complex competitive programming domains.
Following~\citet{islam2024mapcoder}, the problem set size of HumanEval, MBPP, APPS, xCodeEval, and CodeContests are 164, 397, 150, 106, 156, respectively.
We also evaluate on SWE-Bench Verified, Terminal-Bench, and TravelPlanner.

\subsection{Calibration Details and Setup}

\begin{algorithm}[tb]
\caption{Model Adaptive Reflective Prompt Evolution (MAP-RPE) for Model Calibration}
\label{alg:prompt-evolving}
\textbf{Input:} 
Task dataset $S_i$, target model $M_t$, source model $M_s$ (optional), 
prompt database $\mathcal{B}$ with $K$ islands, 
reflection model $\mathcal{M}_{\textrm{ref}}$, 
maximum global iterations $G$, 
local evolution steps $L$ per calibration question,
evaluation metrics $\mathcal{E}$ (e.g., performance, behavioral score).\\
\textbf{Output:} Optimized prompt $p^*_{M_t, S_i}$\\[0.5em]
\textbf{Initialization:} 
Load dataset $S_i$ and construct the initial template $p_0 \leftarrow p_{M_s, S_i}$. 
For each sample $x \in S_i$, get response and store in $\mathcal{B}$. 
Set global best template $p^* \leftarrow p_0$.\\[0.5em]
\For{$g = 1$ \textbf{to} $G$}{
    \For{each calibration instance $x_j \in D_i$}{
        Construct prompt $p_g(x_j)$ from current best template and query $M_t$ to obtain response $r_g$.\\
        Compute metrics $(a_g, b_g) \leftarrow \mathcal{E}(x_j, r_g)$ where $a_t$ denotes task performance and $b_t$ denotes behavioral consistency.\\[0.25em]
        \If{$a_t == Solved$}{
            \textbf{continue} to next instance (solved).}
        \Else{
            Add $\langle p_g, (a_g,b_g)\rangle$ to database $\mathcal{B}$.\\[0.25em]
            \textbf{for} $l = 1$ \textbf{to} $L$ \textbf{do}\\
            \hspace{1em} Select parent prompt $p_{\textrm{parent}}$ and inspirations from current island $\mathcal{B}_k$.\\
            \hspace{1em} Build reflective query $q_l$ with task description, previous template and evaluation feedback.\\
            \hspace{1em} Generate rewritten prompt $p_{\textrm{child}} \leftarrow \mathcal{M}_{\textrm{ref}}(q_l)$.\\
            \hspace{1em} Evaluate $\mathcal{M}_{\textrm{target}}(p_{\textrm{child}}(x_j))$ and record metrics $\mathcal{E}(p_{\textrm{child}})$.\\
            \hspace{1em} Add $p_{\textrm{child}}$ to $\mathcal{B}_k$; update island generation counter.\\
            \hspace{1em} If migration condition met, migrate top programs between islands.\\[0.25em]
            \textbf{end for}
        }
        Update $p^*$ as the best-performing prompt in $\mathcal{B}$ by combined score:
        \[
        p^* \leftarrow \arg\max_{p \in \mathcal{B}} \lambda \cdot \text{Perf}(p) + (1-\lambda) \cdot \text{Behavior}(p)
        \]
    }
}
\Return $p^*_{M_t, S_i} \leftarrow p^*$
\end{algorithm}

\textbf{Behavioral Score }
To ensure that the evolved prompts generate syntactically valid and safe code, we introduce a behavioral score $b \in [0,1]$ as an auxiliary evaluation signal during evolution. This static heuristic assesses the structural and safety properties of a generated completion. The score is composed of four weighted components:
(1) \textbf{Syntax validity} (0.35): whether the generated code can be parsed without syntax errors;
(2) \textbf{Entry-point definition} (0.35): whether the required function (e.g., \texttt{def <entry\_point>(...):}) is correctly defined;
(3) \textbf{Risk-free patterns} (0.20): absence of insecure or potentially harmful code patterns (e.g., \texttt{exec}, \texttt{eval}, file system calls);
(4) \textbf{No undesirable patterns} (0.10): absence of stylistic or task-irrelevant constructs (e.g., print debugging, hardcoded constants).
The final score is a weighted sum of these components, clamped to $[0,1]$. A higher behavioral score reflects well-structured, compliant, and safe code generations, encouraging stable prompt evolution beyond pure functional correctness.
We set $\lambda=0.8$ for all calibration processes.

\textbf{Alignment Tasks }
For each alignment task~\citep{wei2023magicoder, li2022competition}, we apply Model Adaptive Reflective Prompt Evolution (MAP-RPE) to automatically derive the optimized prompt tailored to the target model. The task dataset $S_i$ corresponds to the given alignment task $S_i$ with $n$ instances. Here $n$ is randomly sampled. 
Each training instance consists of a question–answer pair, and the evaluation metric $\mathcal{E}$ measures the textual similarity between the model’s generation and the ground-truth answer.
We set the number of calibration questions to $n=20$, the maximum number of global evolution iterations to $G=20$, and the local evolution steps per question to $L=10$. The prompt archive size is fixed at 1000, and we use $K=3$ evolutionary islands to balance diversity and stability. The population ratios are configured as $exploitation\_ratio=0.7$, $exploration\_ratio=0.2$, and $elite\_selection\_ratio=0.1$. Migration between islands occurs every 50 iterations with a migration rate of 0.1. We use GPT-5 as the prompt optimizer $\mathcal{M}_{ref}$.

\textbf{Unseen Tasks } For unseen coding benchmarks such as HumanEval and xCodeEval, we directly employ the official evaluation metrics of each benchmark, specifically, the functional correctness score, as the reflective evolution objective. In this setting, the evaluation metric $\mathcal{E}$ measures the exact execution-based correctness of generated code rather than textual similarity, enabling the evolution process to optimize for true functional performance.

\subsection{MapCoder Setup}
MapCoder~\citep{islam2024mapcoder} replicates the human programming cycle through four LLM agents - retrieval, plan, code, and debug. It features an adaptive agent traversal schema to interact among corresponding agents dynamically, iteratively enhancing the generated code by, for example, fixing bugs, while maximizing the usage of the LLM agents. The first agent is the Retrieval Agent, recalls pass $k$ relevant problem-solving instances. The second agent is the Planning Agent, aiming to create a step-by-step plan for the original problem. Next is the Coding Agent, which taks the problem description, and a plan from the Planning Agent as input and translates the corresponding planning into code to solve the problem. Finally, the Debugging Agent utilizes sample I/O from the problem description to rectify bugs in the generated code.  We set the similar problems number $k=5$ and the plan generation number $t=5$ for HumanEval. In this setup, the default prompt is treated as the initial transfer point. While it is not explicitly optimized for the source model, it represents a human-crafted, general-purpose prompt, which we consider as the source prompt in this context.

\subsection{TravelPlanner Setup}
TravelPlanner~\citep{xie2024travelplanner} is a planning benchmark that focues on travel planning. It provides a rich sandbox environment, various tools for accessing nearly four million data records, and 1225 meticulously curated planning intents and reference plans. In order to assess whether agents can perceive, understand, and satisfy various constraints to formulate a feasible plan, it includes three types of constraints: Environment Constraints, Commonsense Constraints and Hard Constraints. 
The evaluation criteria include:
\begin{itemize}
    \item \textbf{Delivery Rate:} This metric assesses whether agents can successfully deliver a final plan within a limited number of steps.
    \item \textbf{Commonsense Constraint Pass Rate (CC Pass Rate):} This metric evaluates whether a language agent can incorporate commonsense into their plan without explicit instructions. 
    \item \textbf{Hard Constraint Pass Rate (HC Pass Rate):} This metric evaluates whether a plan satisfies all explicitly given hard constraints in the query. It aims to test the agents’ ability to adapt their plans to diverse user needs.
    \item \textbf{Final Pass Rate:} This metric represents the proportion of feasible plans that meet all aforementioned constraints among all tested plans. It serves as an indicator of agents’ proficiency in producing plans that meet a practical standard. Note that we use this as the primary metric. 
\end{itemize}

TravelPlanner supports two evaluation modes: two-stage mode and sole-planning mode. The two-stage mode is designed to assess the overall capabilities of agents in both tool use and planning. In contrast, the sole-planning mode focuses exclusively on evaluating the agent’s planning skills. In this setting, human-annotated plans are used to predefine the destination cities, and detailed contextual information, such as restaurants within those cities, is directly provided to the agent. This setup eliminates the need for tool calls, as agents no longer need to gather information from scratch. We conduct experiments under both modes. The planning strategies is Direct prompting. We use GPT-5 as our parsing model to improve the parse performance. We also add some guard functions to make the parse more robust.

\section{More Experimental Results}

\subsection{Coding Benchmarks}
\label{app:results:coding}

\begin{table*}
\centering
\caption{Pass@1 Accuracy on several code generation datasets when switching the original model \texttt{GPT-4o} to the target models, including \texttt{Qwen3-32B}, \texttt{Llama3.1-8B-Instruct} and \texttt{Gemma3-27B-it}. Direct Transfer refers to apply the source prompt to the target model directly. Average denotes the mean accuracy across the five code benchmarks, and higher values indicate better performance. The best transfer results are highlighted in \textbf{Bold}.} 
\resizebox{0.9\textwidth}{!}{
\begin{tabular}{c|ccccc|c}
\toprule
 \multicolumn{1}{c}{\textbf{Method}} & \multicolumn{1}{c}{\textbf{HumanEval}}  & \multicolumn{1}{c}{\textbf{MBPP}} & \multicolumn{1}{c}{\textbf{APPS}}    & \multicolumn{1}{c}{\textbf{xCodeEval}} & \multicolumn{1}{c}{\textbf{CodeContests}} & \multicolumn{1}{c}{\textbf{Average}} \\ 
\midrule
Source Model: GPT-4o & 91.10 & 79.80 & 12.00 & 37.03 & 5.45 & 45.08 \\
\midrule
\multicolumn{7}{c}{\textbf{Target Model: Qwen3-32B}} \\
\midrule
Direct Transfer & 72.97  & \textbf{69.10} & 14.45 & 17.61 &13.94 & 37.61 \\
GPT-5 Optimizer & 74.39 &68.85 & 12.00 &  19.50 & \textbf{19.80} & 38.91 \\
\rowcolor{gray!30}  \method (Ours) & \textbf{78.66} & 67.25 & \textbf{16.67} & \textbf{21.70} & 18.79 & \textbf{40.61}\\
\midrule
\multicolumn{7}{c}{\textbf{Target Model: Llama3.1-8B-Instruct}} \\
\midrule
Direct Transfer & 50.60 &50.88 & 2.00 & 6.60 & 7.47 & 23.51 \\
GPT-5 Optimizer & 62.20 & 65.41 & \textbf{5.00}& \textbf{11.32} & 7.27 & \textbf{30.24} \\
\rowcolor{gray!30}  \method (Ours) & \textbf{64.02} & \textbf{66.25} & 1.78 & 9.43 & \textbf{7.68} & 29.83 \\
\midrule
\multicolumn{7}{c}{\textbf{Target Model: Gemma3-27B-it}} \\
\midrule
Direct Transfer & 88.00 & \textbf{79.01} & 9.78 & 22.64 & \textbf{21.21} & 44.13 \\
GPT-5 Optimizer & 87.20 & 78.34 & 14.89 & \textbf{24.80} & 18.99  & 44.84 \\
\rowcolor{gray!30}  \method (Ours) & \textbf{88.01} & 78.59 & \textbf{17.92} & 23.58 & 16.97 & \textbf{45.01}  \\
\bottomrule
\end{tabular}
}
\label{tab:app-gpt4o-others}
\end{table*}

Table~\ref{tab:app-gpt4o-others} presents how well the initial \texttt{GPT-4o} prompt transfer to several target models, \texttt{Qwen3-32B}, \texttt{Llama3.1-8B-Instruct} and \texttt{Gemma3-27B-it} in multiple coding benchmarks.
Across all target models, our method consistently achieves the highest or near-highest accuracy, demonstrating strong generalization of prompt knowledge transfer. For \texttt{Qwen3-32B}, our method improves the average accuracy from 37.61\% (Direct Transfer) to 40.61\%, outperforming GPT-5 Optimizer. For \texttt{Llama3.1-8B-Instruct}, the GPT-5 Optimizer attains the highest average (30.24\%), but our method still surpasses Direct Transfer by a large margin and stays close to the highest average, highlighting reliable transfer even to a small target model. For \texttt{Gemma3-27B-it}, our method achieves the best average score (45.01\%), slightly improving over both baselines while remaining competitive on HumanEval and APPS. Overall, the results indicate that PromptBridge maintains or improves performance relative to Direct Transfer across various model families, validating that model-adaptive prompt transfer reduces degradation when switching to a different LLM.

\subsection{SWE-Bench}
\label{app:results:swe}

\begin{table*}% 'r' for right, 'l' for left
    \centering
    \caption{\textbf{Results on SWE-Bench Verified.} We use \texttt{o4-mini} as the source model and evaluate transferability on \texttt{o3} and \texttt{Llama-3.1-70B-Instruct} as target models. 
    Original denotes the baseline performance using the default prompt template from mini-SWE-agent. 
    Direct Transfer refers to directly applying the optimized prompt from o4-mini to the other models without further adaptation. Relative gain refers to the percentage improvement of PromptBridge over the direct transfer baseline.}
    \label{tab:app-swe-bench}
    \resizebox{0.95\textwidth}{!}{
    \begin{tabular}{l|l|cc}
        \toprule
        & \textbf{Method} & \textbf{Accuracy} & \textbf{Relative Gains}\\
        \midrule
        \multirow{2}{*}{\textbf{\begin{tabular}[c]{@{}c@{}}Source Model: o4-mini\end{tabular}}}
        & Original & 38.60\% & -  \\
        & Optimized & 42.20\% & -\\
        \midrule
       \multirow{3}{*}{\textbf{\begin{tabular}[c]{@{}c@{}}Target Model: o3\end{tabular}}}  
       & Original & 32.00\% & -  \\
       & Direct Transfer (from Optimized) & 33.40\% & 0.00\%\\
       & \method (Ours)& 46.00\% & 27.39\% \\
       \midrule
       \multirow{3}{*}{\textbf{\begin{tabular}[c]{@{}c@{}}Target Model: Llama3.1-70B-Instruct\end{tabular}}}  
       & Original & 6.60\% & -\\
       & Direct Transfer (from Optimized) & 7.60\% & 0.00\% \\
       & \method (Ours) & 8.80\% & 15.79\%\\
        \bottomrule
    \end{tabular}
    }
\end{table*}

\autoref{tab:app-swe-bench} presents the transfer results on SWE-Bench Verified. \method consistently improves performance over direct transfer, achieving relative gains of 27.39\% for \texttt{o3} model and 15.79\% for \texttt{Llama-3.1-70B-Instruct} model, demonstrating its effectiveness in adapting optimized prompts across models. Here, relative gain refers to the percentage improvement of \method over the direct transfer baseline, computed as $(\text{Accuracy}_{\textrm{PromptBridge}} - \text{Accuracy}_{\textrm{Direct}}) / \text{Accuracy}_{\textrm{Direct}} \times 100\%$.
The optimized prompt for \texttt{o4-mini} yields a 9\% improvement compared with the default prompt. However, directly transferring this optimized prompt to other models does not necessarily achieve the best performance, indicating that prompts finely tuned for a specific source model may not generalize optimally to different target models. \textbf{This phenomenon reveals the existence of model drift and underscores the necessity of model-adaptive prompt transfer methods to maintain robust cross-model performance.}

\subsection{Terminal-Bench}
\label{app:results:terminal}

\begin{wrapfigure}{r}{0.4\textwidth}
    \centering
    \includegraphics[width=0.4\textwidth]{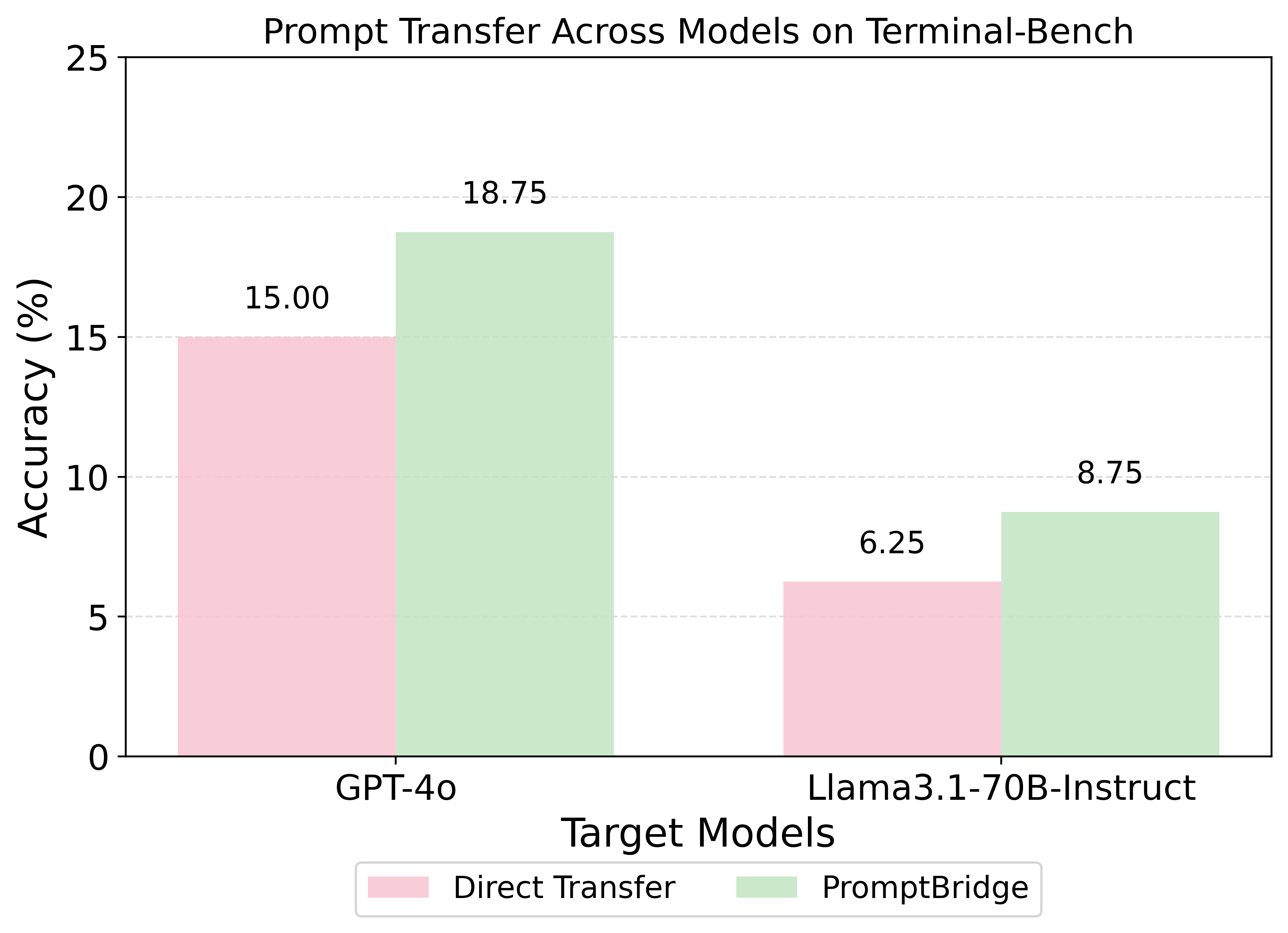}
    \vspace{-15pt}
    \caption{\textbf{Results on Terminal-Bench.} }
    \label{fig:terminal}
\vspace{-10pt}
\end{wrapfigure}

\autoref{fig:terminal} also presents the transfer results on Terminal-Bench. We use \texttt{o3} as the source model and evaluate transferability of the default prompt on \texttt{GPT-4o} and \texttt{Llama-3.1-70B-Instruct} as target models. The source model achieves an accuracy of 36.25\%. 

When transferred to the target models, performance decreases due to differences in model capability and reasoning style. Nevertheless, \method consistently outperforms the direct transfer baseline, achieving relative gains of 25\% on \texttt{GPT-4o} and 40\% on \texttt{Llama-3.1-70B-Instruct}. These results confirm that cross-model prompt transfer through \method effectively mitigates the prompt–model mismatch and enhances transfer robustness across heterogeneous model families.
Note that the prompt used in the direct transfer corresponds to the default prompt employed by the Terminus agent for this figure. Remarkably, \method also improves over these default, non–model-specific optimized prompts, demonstrating its ability to generalize beyond carefully optimized prompts. 

\subsection{TravelPlanner}
\label{app:results:travelplanner}

Beyond the two-stage mode, we also evaluate the sole-planning mode, in which the system reduces to a single-LLM setting focused solely on the planning task. In~\autoref{tab:travelplanner-sole-plan}, the source model is \texttt{GPT-4o}, and the target models are \texttt{o3} and \texttt{Llama3.1-70B-Instruct}. Using the default prompt leads to pass rates of 2.77\% for \texttt{GPT-4o}, 1.67\% for \texttt{o3} and 0.56\% for \texttt{Llama3.1-70B-Instruct}, respectively. 
In this configuration, only the planner prompt is modified while the rest of the system remains fixed.
\method achieves consistent gains over direct transfer and clearly surpasses the default prompt configuration.
Interestingly, the improvements are more pronounced in the sole-planning mode than in the two-stage mode, likely because the planning module is directly responsible for the core reasoning and decision-making process. Thus, adapting its prompt has a more immediate and amplified effect on overall task success with the human-annotated plans.

\begin{table*}
\centering
\caption{Results on TravelPlanner validation set under Sole Planing Mode.} 
\resizebox{\textwidth}{!}{
\begin{tabular}{c|c|cc|cc|c}
\toprule
\textbf{Metric} & \multicolumn{1}{c}{\textbf{Final Pass Rate}}& \multicolumn{2}{c}{\textbf{CC Pass Rate}}& \multicolumn{2}{c}{\textbf{HC Pass Rate}}  & \multicolumn{1}{c}{\textbf{Delivery Rate}}   \\ 
\midrule
\multicolumn{1}{c}{\textbf{Method}} &   &\multicolumn{1}{c}{\textbf{Micro}}    & \multicolumn{1}{c|}{\textbf{Macro}} &\multicolumn{1}{c}{\textbf{Micro}}    & \multicolumn{1}{c|}{\textbf{Macro}} & \\ 
\midrule
Source Model: GPT-4o - Original & 2.77 & \textbf{78.75} & \textbf{19.44} & 30.0 & 13.89 & \textbf{97.22}  \\
% Source Model: GPT-4o - GPT5 & \textbf{3.33} & 43.47 & 10.56 & 25.0 & 13.33 & 52.76 \\
Source Model: GPT-4o - Optimized & \textbf{4.44} & 59.65 & 17.22 & \textbf{32.14} & \textbf{18.89} & 72.22 \\
\midrule
\multicolumn{7}{c}{\textbf{Target Model: o3}} \\
\midrule
Original Prompt  & 1.67 & 54.79 & 2.78 & 1.19 & 1.67 & 93.33 \\
Frozen Direct Transfer & 3.33 & 43.47 & 10.56 & 25.0 & 13.33 & 52.67 \\
\method  & \textbf{7.22} & \textbf{84.10} & \textbf{28.89} & \textbf{36.43} & \textbf{21.67} & \textbf{100.0} \\
\midrule
\multicolumn{7}{c}{\textbf{Target Model: Llama3.1-70B-Instruct}} \\
\midrule
Original Prompt & 0.56 & 40.21 & 1.67 & 0.48 & 1.11 & 67.78 \\
Frozen Direct Transfer  & 1.67 & 54.79 & 10.56 & \textbf{27.14} & \textbf{17.22} & \textbf{70.0} \\
\method & \textbf{3.33} & \textbf{55.14} & \textbf{13.33} & 19.76 & 13.33 & 69.44\\
\bottomrule
\end{tabular}
}
\label{tab:travelplanner-sole-plan}
\end{table*}

\subsection{Consistency Analysis}
\label{app:results:consistency}
\begin{figure}[t]
    \centering
    \includegraphics[width=\textwidth]{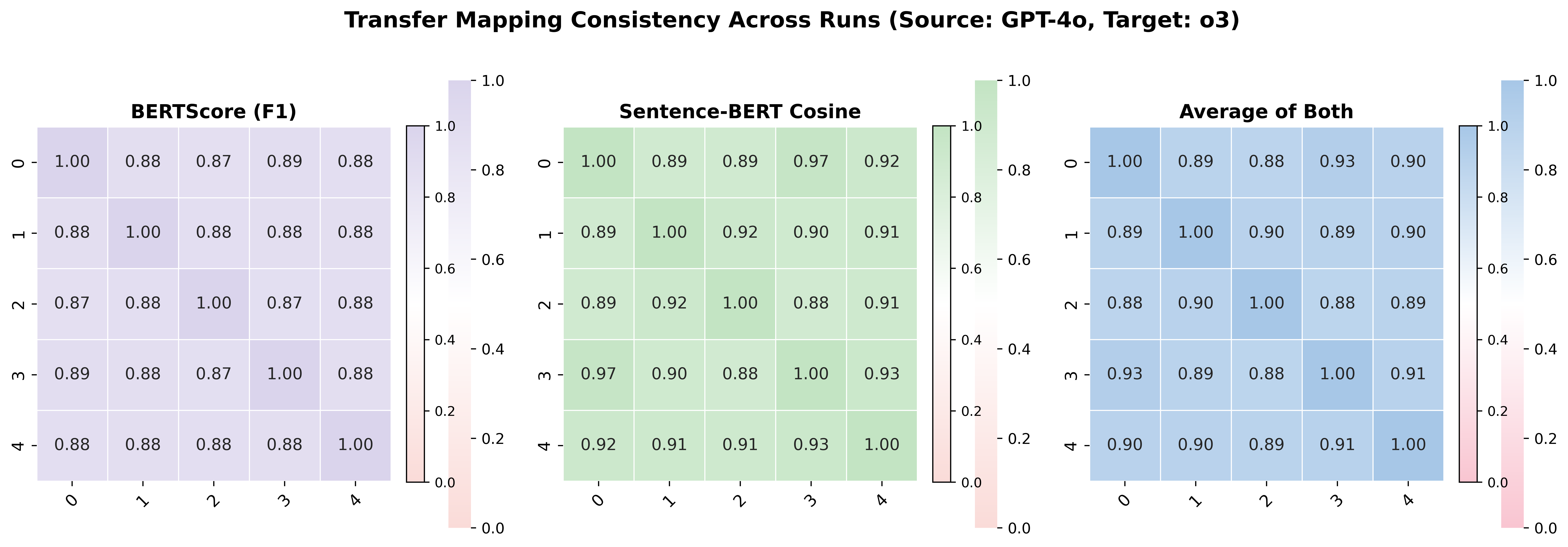}
    \vspace{1em} % adjust vertical space between figures
    \includegraphics[width=\textwidth]{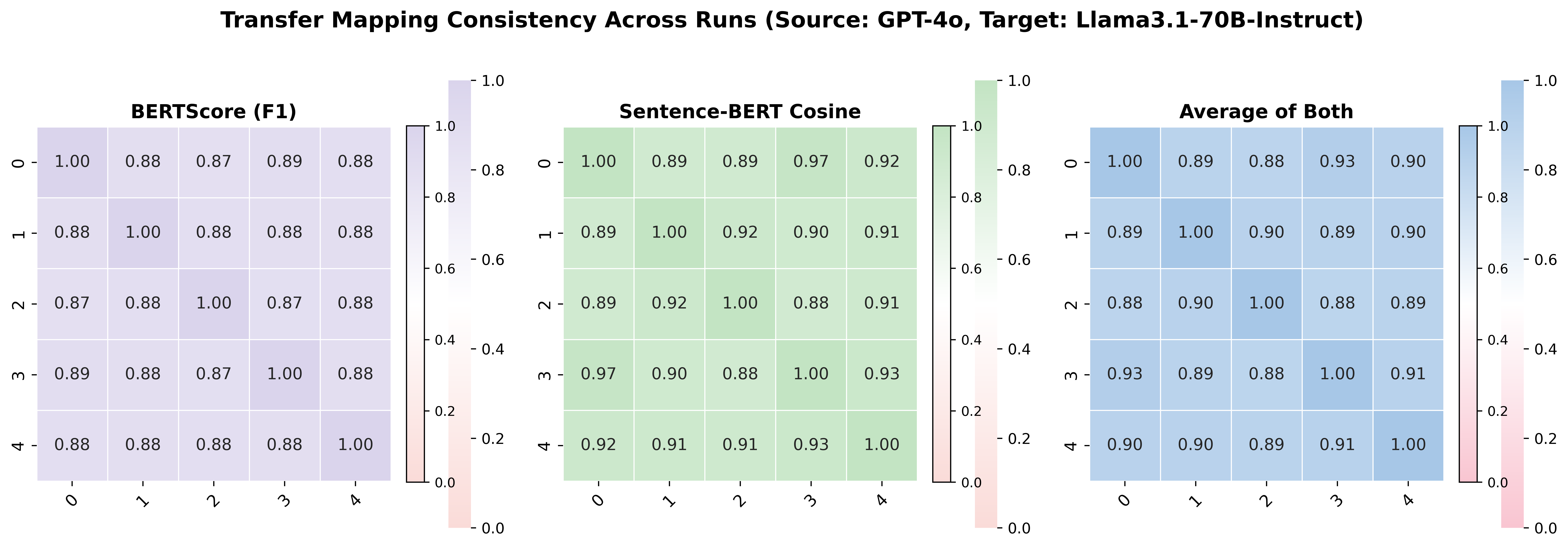}
    \caption{\textbf{Transfer Mapping Consistency Across Runs.} 
Each heatmap shows the pairwise similarity of transfer mappings obtained from five independent runs under identical configurations. 
The top panel reports consistency when transferring from \texttt{GPT-4o} to \texttt{o3}, while the bottom panel reports transfer from \texttt{GPT-4o} to \texttt{Llama-3.1-70B-Instruct}. Higher off-diagonal values indicate stronger stability of the transfer effects across repeated runs. 
}
    \label{fig:consistency-summary}
\end{figure}

\begin{figure}[t]
    \centering
    \includegraphics[width=\textwidth]{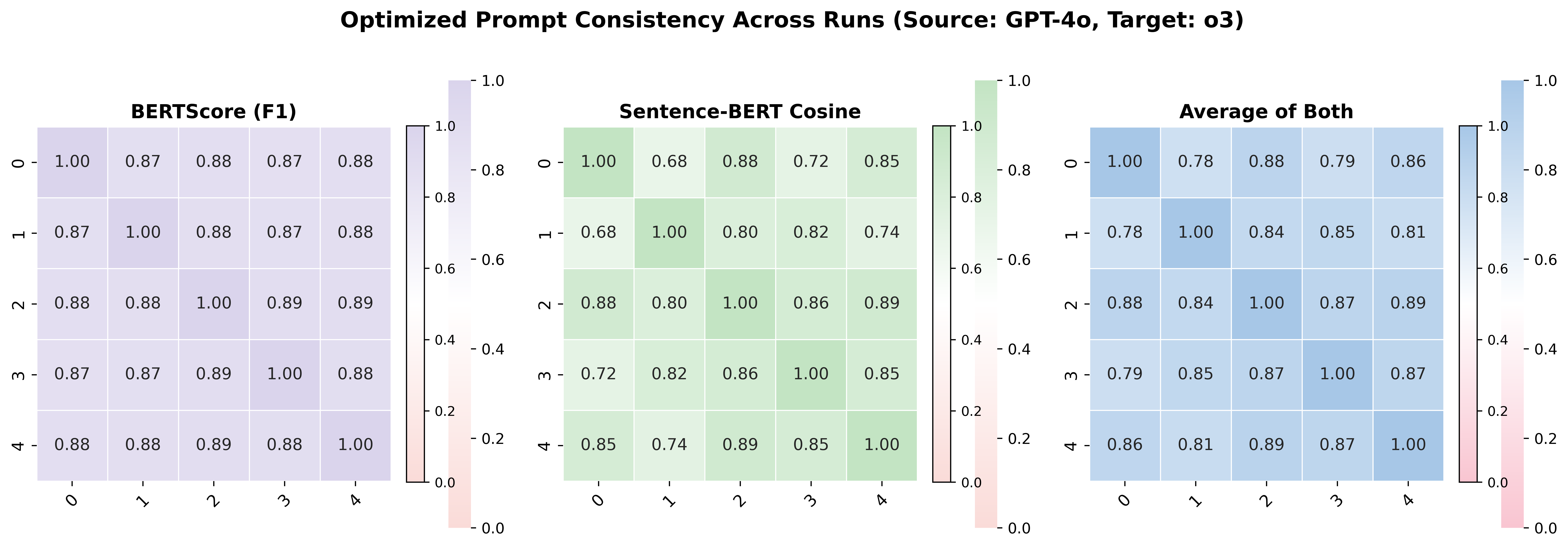}
    \vspace{1em} % adjust vertical space between figures
    \includegraphics[width=\textwidth]{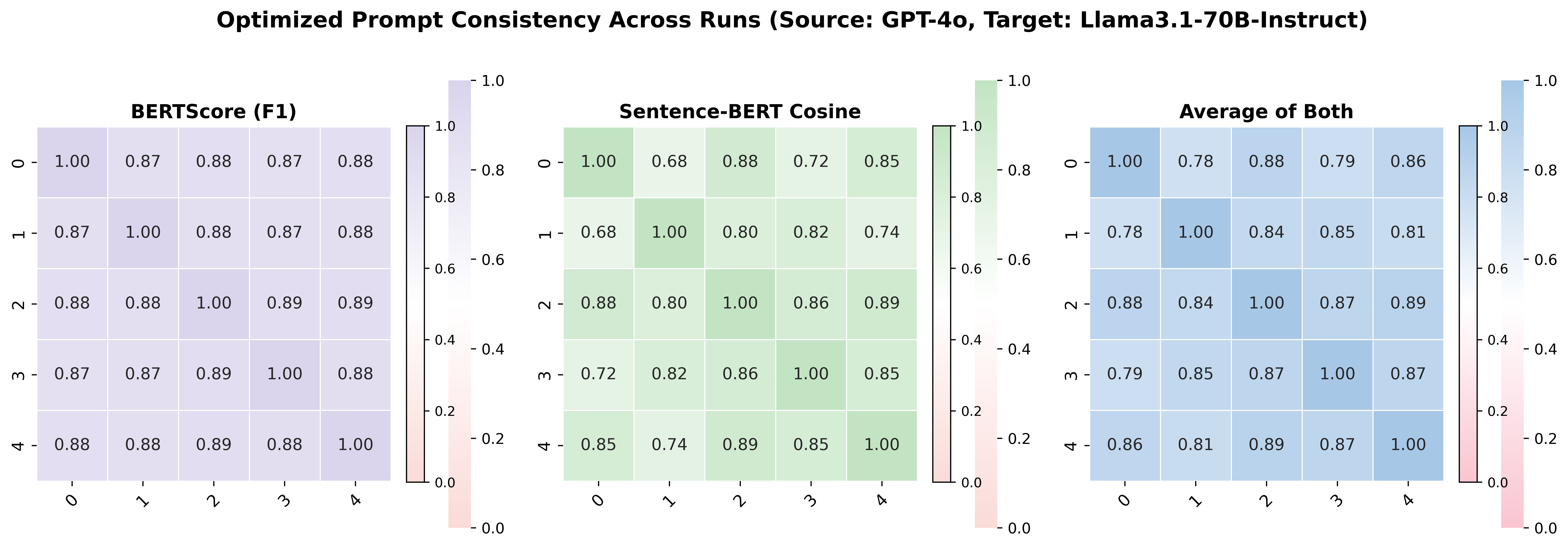}
    \caption{\textbf{Optimized Prompt Consistency Across Runs.}
Each heatmap visualizes the pairwise similarity among optimized prompts obtained from five independent runs under identical configurations. 
The top panel corresponds to transfer from \texttt{GPT-4o} to \texttt{o3}, and the bottom panel corresponds to transfer from \texttt{GPT-4o} to \texttt{Llama-3.1-70B-Instruct}. 
Diagonal entries denote perfect self-similarity (1.0), while higher off-diagonal values reflect greater stability of the optimized prompt across runs.}
    \label{fig:consistency-prompt}
\end{figure}
To evaluate the stability of PromptBridge’s cross-model adaptation, we assess the similarity of the generated transfer mapping and the optimized prompts across multiple independent runs with identical configurations. Since LLM-based systems may introduce stochastic variation in generation, it is important to quantify whether the learned transfer mapping remain semantically consistent across runs. The same semantic stability should also manifest in the optimized prompts. We conduct the experiments when transferring \texttt{GPT-4o} to \texttt{o3}, and \texttt{Llama3.1-70B-Instruct}. Specifically, we compute pairwise similarity scores between transfer mapping generated from $N=5$ independent runs using semantic metrics. We adopt two complementary embedding-based similarity measures: 
\begin{itemize}
    \item \textbf{BERTScore (F1)~\citep{zhang2019bertscore}:} measures semantic alignment by comparing contextual embeddings of tokens between two texts using a pre-trained language model (BERT). It captures fine-grained word-level correspondence and yields higher values when two sentences convey equivalent meaning, even with different surface forms.
    \item \textbf{Sentence-BERT Cosine Similarity~\citep{reimers2019sentence}:} represents each text as a holistic sentence embedding derived from the all-mpnet-base-v2 model\footnote{https://huggingface.co/sentence-transformers/all-mpnet-base-v2}, and computes the cosine similarity between embeddings. This provides a global semantic similarity measure at the sentence or paragraph level, complementing the token-level sensitivity of BERTScore.
\end{itemize}

We compute pairwise similarities using both metrics, and report their averaged results. 
Diagonal entries represent self-similarity (1.0). Higher scores indicate greater semantic consistency and thus higher robustness of PromptBridge to sampling randomness.

\paragraph{Transfer Mapping Consistency Analysis }
To assess whether the learned transfer mapping remain stable across repeated executions, we conduct five independent runs of PromptBridge under identical configurations and compute pairwise similarity between the generated transfer mappings. As shown in \autoref{fig:consistency-summary}, both BERTScore and Sentence-BERT cosine similarity exhibit consistently high off-diagonal values (mostly between 0.85–0.90), suggesting strong semantic alignment across runs. The \texttt{GPT-4o} → \texttt{o3} transfer achieves an average cross-run consistency of approximately 0.87, while the \texttt{GPT-4o} → \texttt{Llama-3.1-70B-Instruct} transfer demonstrates slightly higher stability, with average similarity around 0.89. The small variance across runs implies that PromptBridge yields reproducible transfer mappings even under inherent stochasticity in LLM generations. This confirms that the optimization process captures a stable underlying adaptation pattern rather than overfitting to random noise.

\paragraph{Optimized Prompt Consistency Analysis.}
To further examine whether the optimized prompts produced by PromptBridge remain stable across repeated executions, we perform five independent optimization runs using identical settings and compute the pairwise similarity among the resulting prompts. 
As illustrated in \autoref{fig:consistency-prompt}, 
for the \texttt{GPT-4o} → \texttt{o3} transfer, the average similarity across runs remains around 0.82, reflecting a stable prompt search behavior. 
In contrast, the \texttt{GPT-4o} → \texttt{Llama-3.1-70B-Instruct} transfer shows a wider variation (around 0.78–0.89 for average similarity). 
Overall, the results demonstrate that while the optimized prompts are not identical across runs, they remain semantically coherent, indicating that PromptBridge consistently converges toward functionally equivalent solutions rather than overfitting to specific sampling randomness.

\subsection{Ablation Study}
\paragraph{Ablation Study on the Mapping Extractor and Adapter Model.}
\autoref{tab:ablation-model} reveals clear trends regarding how the choice of Mapping Extractor and Adapter Model influences transfer quality. When both components are implemented using \texttt{GPT-5}, PromptBridge achieves the strongest performance, consistent with the fact that \texttt{GPT-5} is the most capable model available. In contrast, using a weaker model such as \texttt{Qwen3-32B} for either component leads to the lowest accuracy, especially since it is neither the source nor the target model and it is weaker than others.
For the Mapping Extractor, when the Adapter Model is fixed to \texttt{GPT-5}, \texttt{GPT-5} as the mapping yields the best performance, followed by \texttt{Llama3.1-70B-Instruct}, and then \texttt{GPT-4o}. The competitive performance of \texttt{Llama3.1-70B-Instruct} in this setting may because it matches the target model, making it naturally aligned with the target prompt distribution.
For the Adapter Model, we observe a consistent improvement whenever \texttt{GPT-5} is used as the adapter: holding the Mapping Extractor fixed, configurations with \texttt{GPT-5} as the adapter always outperform those using weaker adapter models. This suggests that a powerful adapter model is particularly important for accurately synthesizing the target-compatible prompt once the mapping has been learned.

\begin{table*}
\centering
\caption{\textbf{Ablation study on the Mapping Extractor and Adapter Model.}
Using \texttt{GPT-4o} as the source model and \texttt{Llama3.1-70B-Instruct} as the target, we vary the Mapping Extractor and Adapter Model within PromptBridge to assess their impact on transfer performance. Direct Transfer serves as the baseline without any mapping or adaptation. Average reports the mean accuracy across HumanEval and xCodeEval. Best results are highlighted in \textbf{bold}, and second-best results are \underline{underlined}.
} 
\resizebox{0.8\textwidth}{!}{
\begin{tabular}{cc|cc|c}
\toprule
\multicolumn{1}{c}{\textbf{Mapping Extractor}} & \multicolumn{1}{c|}{\textbf{Adapter Model}}& \multicolumn{1}{c}{\textbf{HumanEval}}  & \multicolumn{1}{c|}{\textbf{xCodeEval}} & \multicolumn{1}{c}{\textbf{Average}}  \\ 
\midrule
\multicolumn{2}{c|}{Direct Transfer} & 68.70 & 21.38 & 45.04 \\
\midrule 
GPT-5 & GPT-5 &  \underline{79.88} & \textbf{24.53} & \textbf{52.21} \\
GPT-4o & GPT-5 & 76.83 & \underline{23.58} & 50.21 \\
GPT-4o & GPT-4o & 76.22 & 19.81 & 48.02 \\
Qwen3-32B & GPT-5 & 78.66 & 21.70 & 50.18 \\
Qwen3-32B & Qwen3-32B  & 51.22 & 22.64 & 36.93\\
Llama3.1-70B-Instruct & GPT-5 & \textbf{81.10} & 21.70 & \underline{51.40} \\
Llama3.1-70B-Instruct  & Llama3.1-70B-Instruct  & 78.66 &  16.98 & 47.82\\
\bottomrule
\end{tabular}
}
\label{tab:ablation-model}
\end{table*}

\section{Discussions}
\paragraph{Discussions on Migration Effort.} 
When migrating to a new model, whether for improved performance, reduced cost, or enhanced privacy—deriving effective prompts for the target model can be costly and labor-intensive.
Our framework addresses this challenge through \method,
a training-free prompt transfer framework that utilizes the MAP-RPE to calibrate model-specific prompt adaptation on a small set of alignment tasks. Once this calibration is complete, the learned transfer effects can be efficiently propagated to unseen tasks, thereby minimizing task-specific optimization effort and enabling scalable cross-model deployment.

A closely related work, MAPO~\citep{chen2024mapo}, also performs model-adaptive prompt optimization but relies on a resource-intensive pipeline involving dataset construction, supervised fine-tuning, and reinforcement learning. In contrast, our approach is entirely training-free and lightweight, yet achieves generalizable adaptation across diverse models and tasks.
By integrating MAP-RPE for model calibration and PromptBridge for knowledge transfer, we provide a practical and scalable solution for reducing the migration cost in single-model, single-agent and multi-agent systems.

\paragraph{Real-world Application.}
Model Drifting is critical in various scenarios where it is necessary to switch one core LLM to another. We provide three situations as follow:
1) \textbf{Powerful Model for better performance.} Latest advanced model can largely improve the performance of the LLM system. Thus the existing system may face the need to upgrade their own core LLM to improve the performance or keep upgraded with the best service.
2) \textbf{Lightweight Model for lower cost.} To reduce API expenses or enable on-premise deployment, one may replace the source LLM with a smaller or more cost-efficient model. Such transitions must preserve performance despite the reduced capacity of the target model.
3) \textbf{Open-source Model for control and privacy.} Organizations may migrate from proprietary APIs to open-source models to gain stronger privacy guarantees, greater customizability, or the ability to further train the model (e.g., via instruction tuning or reinforcement learning).
In all these scenarios, the transferred prompt $\widehat{P}_{M_t, T}$ should ensure that the target model's performance remains comparable to that of the source model. Moreover, for any unseen task $T_j \in T$, the learned transfer function $\mathcal{T}$ should reliably generate an effective target-model prompt. As LLM systems become increasingly complex, it is crucial that the prompt transfer method remains low-cost, given the growing number of models and downstream tasks.

\paragraph{Limitations.} Our transfer effects are learned from standard tasks and model families; coverage may diminish for niche domains or rapidly updated model APIs. Prompt evolution incurs nontrivial compute. \method currently focuses on optimizing instructions alone, omitting exemplar or few-shot demonstration optimization.

\section{Prompt Template}
\label{app:prompt_template}
Owing to space constraints, the complete set of prompt templates will be provided in the released codebase.

\subsection{PromptBridge}

\label{app:prompt:method}
\begin{tcolorbox}[
    enhanced,
    breakable,
    colback=mintback,
    colframe=mintframe,
    colbacktitle=mintframe,
    coltitle=white,
    boxrule=1.5pt,
    arc=1mm,
    left=10pt,
    right=10pt,
    top=10pt,
    bottom=10pt,
    fonttitle=\bfseries\large,
    title={Mapping Extractor Prompt. },
    attach boxed title to top left={yshift=-2mm}
]
\subsection*{System Prompt}
You are a helpful assistant that summarizes the difference of prompts.

\subsection*{User Prompt}
Below are \{m\} examples of the source prompts and target prompts, along with their dataset and information on the dataset.

Source Prompt \{1\}: \{source\_prompts[0]\}

Target Prompt \{1\}: \{target\_prompts[0]\}

Dataset: \{infos[0]\}

......

Source Prompt \{m\}: \{source\_prompts[m-1]\}

Target Prompt \{m\}: \{target\_prompts[m-1]\}

Dataset: \{infos[m-1]\}

Please summarize the common prompt difference of the source prompts to the target prompts, also considering the dataset and information.
\end{tcolorbox}

\begin{tcolorbox}[
    enhanced,
    breakable,
    colback=mintback,
    colframe=mintframe,
    colbacktitle=mintframe,
    coltitle=white,
    boxrule=1.5pt,
    arc=1mm,
    left=10pt,
    right=10pt,
    top=10pt,
    bottom=10pt,
    fonttitle=\bfseries\large,
    title={Adapter Prompt for Coding Benchmark. },
    attach boxed title to top left={yshift=-2mm}
]
Your task is to apply the transfer effects from the source prompt to generate a new target prompt.

The transfer effects were derived from the standard coding dataset.

You must now generate a prompt for the unseen dataset that incorporates these transfer effects.

Begin from the Original Prompt provided below.

\#\# ====== Original Prompt Starts ========

\#\# Original Prompt: \{Source Prompt\}

\#\# ====== Original Prompt Ends ========

\vspace{0.5em}

\#\# Transfer Effects Summary:

\{summary\}

\vspace{0.5em}

\#\# Task:

Apply the above transfer effects summary to the Original Prompt designed for \{source\_model\}.

Generate a new prompt that is:

- Adapted for the \{target\_model\} model,

- Grounded in the transfer effects summary,

- Suitable for eliciting higher-quality responses on the coding datasets, such as HumanEval and xCodeEval.

\vspace{0.5em}

Optimized Prompt: 
\end{tcolorbox}

\begin{tcolorbox}[
    enhanced,
    breakable,
    colback=mintback,
    colframe=mintframe,
    colbacktitle=mintframe,
    coltitle=white,
    boxrule=1.5pt,
    arc=1mm,
    left=10pt,
    right=10pt,
    top=10pt,
    bottom=10pt,
    fonttitle=\bfseries\large,
    title={Adapter Prompt for SWE-Bench. },
    attach boxed title to top left={yshift=-2mm}
]
Your task is to generate a new target prompt by applying the specified transfer effects to the Original Prompt. 

These transfer effects were derived from a standard coding dataset and must now be adapted for SWE-Bench.  

\vspace{0.5em}

The new prompt should:  

- Begin from the provided Original Prompt.  

- Incorporate the transfer effects summary faithfully.  

- Be adapted for the \{target\_model\} model.  

- Remain concise and preserve the original meaning.  

- Improve suitability for eliciting high-quality responses on complex agent benchmarks such as SWE-Bench. 

\vspace{0.5em}

\#\# ====== Original Prompt ======  

\{original\_prompt\}  

\#\# ====== End Original Prompt ======  

\vspace{0.5em}

\#\# ====== Transfer Effects Summary ======  

\{summary\}

\#\# ====== End Transfer Effects Summary ======  

\vspace{0.5em}

**Task:**  

Apply the transfer effects summary to the Original Prompt optimized for \{source\_model\} and produce an optimized prompt for \{target\_model\}. 

\vspace{0.5em}

Optimized Prompt: 
\end{tcolorbox}

\begin{tcolorbox}[
    enhanced,
    breakable,
    colback=mintback,
    colframe=mintframe,
    colbacktitle=mintframe,
    coltitle=white,
    boxrule=1.5pt,
    arc=1mm,
    left=10pt,
    right=10pt,
    top=10pt,
    bottom=10pt,
    fonttitle=\bfseries\large,
    title={Adapter Prompt for Terminal-Bench. },
    attach boxed title to top left={yshift=-2mm}
]
Your task is to generate a new target prompt by applying the specified transfer effects to the Original Prompt.

These transfer effects were derived from a standard coding dataset and must now be adapted for Terminal Bench.   

\vspace{0.5em}

The new prompt should:  

- Begin from the provided Original Prompt.  

- Incorporate the transfer effects summary faithfully.  

- Be adapted for the \{target\_model\} model.  

- Remain concise and preserve the original meaning.  

- Improve suitability for eliciting high-quality responses on complex agent benchmarks such as Terminal Bench.  

\vspace{0.5em}

\#\# ====== Original Prompt ======  

\{original\_prompt\}  

\#\# ====== End Original Prompt ======  

\vspace{0.5em}

\#\# ====== Transfer Effects Summary ======  

\{summary\}

\#\# ====== End Transfer Effects Summary ======  

\vspace{0.5em}

**Task:**  

Apply the transfer effects summary to the Original Prompt optimized for \{source\_model\} and produce an optimized prompt for \{target\_model\}. 

\vspace{0.5em}

Optimized Prompt: 
\end{tcolorbox}

\begin{tcolorbox}[
    enhanced,
    breakable,
    colback=mintback,
    colframe=mintframe,
    colbacktitle=mintframe,
    coltitle=white,
    boxrule=1.5pt,
    arc=1mm,
    left=10pt,
    right=10pt,
    top=10pt,
    bottom=10pt,
    fonttitle=\bfseries\large,
    title={Adapter Prompt for TravelPlanner. },
    attach boxed title to top left={yshift=-2mm}
]
Your task is to generate a new target prompt by applying the specified transfer effects to the Original Prompt.  

These transfer effects were derived from several standard datasets and must now be adapted for a planning agent benchmark.  

\vspace{0.5em}

The new prompt should:  

- Begin from the provided Original Prompt.  

- Incorporate the transfer effects summary faithfully.  

- Be adapted for the \{target\_model\} model.  

- Remain concise and preserve the original meaning.  

- Improve suitability for eliciting high-quality responses on complex agent benchmarks such as TravelPlanner.  

\vspace{0.5em}

\#\# ====== Original Prompt ======  

\{original\_prompt\}  

\#\# ====== End Original Prompt ======  

\vspace{0.5em}

\#\# ====== Transfer Effects Summary ======  

\{summary\}

\#\# ====== End Transfer Effects Summary ======  

\vspace{0.5em}

**Task:**  

Apply the transfer effects summary to the Original Prompt optimized for \{source\_model\} and produce an optimized prompt for \{target\_model\}. 

\vspace{0.5em}

Optimized Prompt: 
\end{tcolorbox}

\subsection{Optimized Prompt for Coding Benchmarks}
\label{app:prompt:coding}
\begin{tcolorbox}[
    enhanced,
    breakable,
    colback=roseback,
    colframe=roseframe,
    colbacktitle=roseframe,
    coltitle=white,
    boxrule=1.5pt,
    arc=1mm,
    left=10pt,
    right=10pt,
    top=10pt,
    bottom=10pt,
    fonttitle=\bfseries\large,
    title={PromptBridge (GPT-4o $\rightarrow$ o3) },
    attach boxed title to top left={yshift=-2mm}
]
You are the o3 model. Read the programming problem below and produce a correct, efficient, and idiomatic Python solution.

\vspace{0.5em}

Problem:
\{Input Coding Problem\}

\vspace{0.5em}

Requirements:

- Implement exactly the API required by the problem:

- Use the exact function name(s) and signature(s) specified or implied by the problem/stub.

- Do not rename functions, change parameter order/names, or alter return types.

- Output format:

   \hspace{1em} - Return only the final Python source code with no explanations, prose, or markdown fences.
    
    \hspace{1em}- Do not include any I/O (no print, input, file reads/writes), no main guard, and no demo/tests.
    
- Code standards:

   \hspace{1em} - Python 3.8+ compatible. Use only the standard library.
    
   \hspace{1em} - Include precise type hints and a concise docstring for each public function:
    
    \hspace{1em} \hspace{1em}   - Describe parameters, return value, key edge cases, and expected complexity.
        
    \hspace{1em}- Prefer iterative solutions unless recursion is clearly appropriate and safe for input sizes.
    
   \hspace{1em} - Avoid global state; write pure functions unless mutation is explicitly required.
    
    \hspace{1em}- Do not modify inputs in place unless the problem explicitly asks for it; otherwise treat inputs as immutable.
    
- Correctness and edge cases:

   \hspace{1em} - Handle typical edge cases that logically apply to the problem: empty inputs, single-element cases, duplicates, ties, negative values, zeros, large values, and unicode/whitespace handling for strings\u2014as relevant to the task.
    
   \hspace{1em} - Maintain determinism: avoid randomness and nondeterministic data structure iteration affecting outputs.
    
   \hspace{1em} - Be careful with integer overflow (use Python\u2019s big integers) and floating-point precision (prefer exact arithmetic or stable comparisons when required).
    
   \hspace{1em} - If multiple valid outputs exist, follow the canonical/most widely accepted choice (e.g., lexicographically smallest) if the problem implies it; otherwise document the chosen tie-break in the docstring and implement consistently.
    
- Efficiency:

  \hspace{1em}  - Choose data structures and algorithms to meet typical constraints for coding benchmarks (aim for optimal or near-optimal time/space).
    
   \hspace{1em} - Avoid unnecessary quadratic behavior on large inputs; justify any inherently superlinear steps by problem requirements.
    
- Ambiguity handling:

   \hspace{1em} - If the problem is underspecified, adopt the most standard interpretation and note any minimal assumptions in the function\u2019s docstring (no extra commentary outside the code).
    
- Validation:

   \hspace{1em} - Ensure the implementation would pass the examples implied by the problem statement.
    
  \hspace{1em}  - Do not raise exceptions unless explicitly required by the specification; prefer returning the specified sentinel/structure.

\vspace{0.5em}

Deliverable:

- Only the complete Python solution code, conforming to the above, with the required function(s) implemented. No extra text.
\end{tcolorbox}

\begin{tcolorbox}[
    enhanced,
    breakable,
    colback=roseback,
    colframe=roseframe,
    colbacktitle=roseframe,
    coltitle=white,
    boxrule=1.5pt,
    arc=1mm,
    left=10pt,
    right=10pt,
    top=10pt,
    bottom=10pt,
    fonttitle=\bfseries\large,
    title={PromptBridge for Debugging Agent in MapCoder (GPT-4o $\rightarrow$ o3) },
    attach boxed title to top left={yshift=-2mm}
]
Task: Assess whether the provided plan will correctly solve the given competitive programming problem under standard coding benchmark constraints (HumanEval, xCodeEval).

\vspace{0.5em}

Inputs:

- Problem: \{Input Coding Problem\}

- Plan: \{Planning\}

\vspace{0.5em}

Instructions:

- Assume Python 3.10, standard library only, deterministic logic (no randomness or external state), and robust handling of all edge cases.

- Evaluate the plan against typical grader expectations:

\hspace{1em}- I/O policy: If the problem requires a specific function signature, the solution must implement exactly that signature with no stdin/stdout. If the problem is script-style, the solution must read from stdin and write to stdout precisely, with no extra output.

\hspace{1em}- Time and space complexity: Confirm the plan is near-optimal for worst-case constraints; flag potential TLE or memory issues.

\hspace{1em}- Correctness: Provide a proof sketch or invariant-based reasoning that the algorithm produces correct results for all cases, including corner cases (empty inputs, extremes, duplicates, negative values, large sizes).

\hspace{1em}- Determinism and stability: Verify tie-breaking rules (e.g., lexicographic/stable ordering) and output formatting are properly handled when unspecified.

\hspace{1em}- Numeric issues: Check integer overflow assumptions, floating-point precision/rounding, and formatting requirements.

\hspace{1em}- Recursion vs iteration: Prefer iterative approaches when recursion depth may exceed Python limits; note stack risks or the need for tail-call avoidance.

\hspace{1em}- Data structures: Validate chosen structures and operations (sorting stability, hashing assumptions, boundary conditions).

\hspace{1em}- I/O formatting: Ensure no extra whitespace/lines, exact printing requirements, and consistent encoding.

- Be concise, precise, and deterministic. Do not include code, examples, or any text outside the required XML.

- Output must be exactly the following XML with two tags. Do not add markdown or any additional text:

\vspace{0.5em}

<root>

<explanation>Discuss whether the given competitive programming problem is solvable by using the above mentioned planning. Address algorithmic correctness, complexity, edge cases, I/O requirements, determinism, and any grader-aligned constraints. Clearly state if the plan is correct, partially correct, or incorrect, and why.</explanation>

<confidence>Confidence score regarding the solvability of the problem. Must be an integer between 0 and 100.</confidence>

</root>"

\end{tcolorbox}

\subsection{Optimized Prompt for SWE-Bench and Terminal-Bench}
\label{app:prompt:swe-terminal}

\begin{tcolorbox}[
    enhanced,
    breakable,
    colback=skyback,
    colframe=skyframe,
    colbacktitle=skyframe,
    coltitle=white,
    boxrule=1.5pt,
    arc=1mm,
    left=10pt,
    right=10pt,
    top=10pt,
    bottom=10pt,
    fonttitle=\bfseries\large,
    title={PromptBridge (o4-mini $\rightarrow$ o3) on SWE-Bench. },
    attach boxed title to top left={yshift=-2mm}
]
<pr\_description>

Consider the following PR description:

\{\{task\}\}

</pr\_description>

\vspace{0.5em}

<instructions>

\# Task Instructions (o3-optimized)

\vspace{0.5em}

\#\# Role and Goal

You are a senior software engineer interacting with a Linux shell, issuing one command at a time. Your goal is to implement changes in non-test files to satisfy the PR description, in a way that is general, performant, and consistent with the codebase.

\vspace{0.5em}

\#\# Key Boundaries

- MODIFY: Regular source files under /testbed (this is your working directory).

- DO NOT MODIFY: Tests, configuration files (pyproject.toml, setup.cfg, etc.).

- Preserve existing public interfaces and function/class signatures unless the PR explicitly requires changes.

- Use only the standard library and existing project dependencies; do not add new external deps.

- Avoid OS-level side effects unless strictly necessary for the fix.

\vspace{0.5em}

\#\# Code Quality and Performance

- Add type hints to new or modified public functions.

- At the top of modified modules, include a concise module docstring: approach and Big-O; at most one explicit assumption if needed.

- Prefer iterative solutions; avoid deep recursion.

- Aim for $O(n)–O(n log n)$; avoid $O(n^2)$ on large inputs unless inherent to the problem.

- Ensure deterministic behavior, stable ordering, and lexicographic tie-breaking if unspecified.

- Do not mutate input objects unless the API expects it or the codebase already relies on it.

- Exact output formatting when producing any script output: no extra prints, robust Unicode handling, avoid unintended scientific notation.

\vspace{0.5em}

\#\# Workflow

1. Analyze the repo to locate relevant code paths and constraints.

2. Create a non-interactive reproducibility script (if helpful) that reads from sys.stdin.buffer and writes to stdout; include a main entry point.

3. Implement fixes in source files, adhering to existing signatures/APIs and code style.

4. Re-run your script to verify the fix.

5. Probe edge cases to ensure robustness.

\vspace{0.5em}

\#\# Command Execution Rules

This is an interactive loop:

- You will think, then issue ONE command; the system executes it; you see the result; then you issue your next command.

- Each response MUST have:

  \hspace{1em}1) A THOUGHT section explaining your reasoning and intent.
  
  \hspace{1em}2) EXACTLY ONE bash code block containing EXACTLY ONE command (or a single compound command chained with \&\& or ||).
  
- Environment state (like cd or env vars) is not persistent across commands; prefix actions accordingly (e.g., cd /testbed \&\& ...).

- Always use non-interactive flags (-y, -f). Avoid interactive editors/tools.

\vspace{0.5em}

Format your responses like this:

\vspace{0.5em}

<format\_example>

THOUGHT: Explain your current reasoning, what you aim to learn/change, and why the single command below is the best next step.

\vspace{0.5em}

```bash

your\_command\_here

```

</format\_example>

\vspace{0.5em}

If you need multiple steps, execute them sequentially across responses, or chain with \&\& or || in a single command.

\vspace{0.5em}

\#\# Useful Command Examples

- Create a new file:

\begin{verbatim}
cat <<'EOF' > script.py
import sys
def main() -> None:
    data = sys.stdin.buffer.read()
   # process and print exact output
    print("ok")
if __name__ == "__main__":
   main()
EOF
\end{verbatim}

- Edit files with sed:

\begin{verbatim}
sed -i 's/old/new/g' filename.py
\end{verbatim}

- View file content:

\begin{verbatim}
nl -ba filename.py | sed -n '10,30p'
\end{verbatim}

\vspace{0.5em}

\#\# Submission

When your fix is complete and you cannot make further progress, submit exactly:

\begin{verbatim}
echo COMPLETE_TASK_AND_SUBMIT_FINAL_OUTPUT && git add -A \\
&& git diff --cached
\end{verbatim}

You cannot continue working after submitting.

</instructions>
\end{tcolorbox}

\begin{tcolorbox}[
    enhanced,
    breakable,
    colback=skyback,
    colframe=skyframe,
    colbacktitle=skyframe,
    coltitle=white,
    boxrule=1.5pt,
    arc=1mm,
    left=10pt,
    right=10pt,
    top=10pt,
    bottom=10pt,
    fonttitle=\bfseries\large,
    title={PromptBridge (GPT-4o $\rightarrow$ o3) on Terminal-Bench. },
    attach boxed title to top left={yshift=-2mm}
]

You control a Linux terminal inside a tmux session. Your job is to solve the task by iteratively sending minimal batches of terminal inputs (shell commands and/or keystrokes) and requesting output only when it is necessary to make correct progress. Respond with a single JSON object that exactly matches the given schema.

\vspace{0.5em}

Follow this loop every time you respond:

1) Analyze the latest terminal output/state concisely and deterministically.

2) Plan the minimal next actions that move directly toward the instruction.

3) Emit the next batch of inputs.

4) Decide if you need to see the resulting output before continuing, and set the need-output flag accordingly.

\vspace{0.5em}

Strict output requirements:

- Output must be a single JSON object matching \{response\_schema\}. Do not add extra fields, comments, or markdown.

- Where the schema collects the terminal inputs (e.g., a list/array of commands), each entry must be one of:

  - A shell command terminated with a newline character: "actual command text
"

  - A keystroke sequence for interactive apps, without a trailing newline (e.g., "q", "ESC", "C-c", ":q!", "j", "k"). Only use newline for the Enter key.
  
- If you open a full-screen/interactive program (less, vim, man, git diff, top, etc.), do not wait for its output. Immediately send the keystrokes needed to navigate/exit, as separate entries, until you return to the shell prompt.

\vspace{0.5em}

Operational constraints and discipline (adapted for complex agent benchmarks):

- Be precise, deterministic, and self-contained. Never use placeholders like <file>; infer real paths from the current state.

- Prefer non-interactive, idempotent commands and flags (sed -i, printf, tee, here-docs with explicit delimiters) over editors. Only use interactive tools when required, and then exit cleanly with keystrokes.

- I/O discipline: only request terminal output when needed for correctness; avoid producing or paginating large outputs unnecessarily; use quiet flags (-q), targeted filters (grep -F/-E, awk), and stable ordering (sort -s) to make parsing reliable.

- Library/execution constraints: use standard shell utilities available in a typical Linux environment; no networking, no background daemons, and no subprocess launches beyond normal shell commands; avoid sudo unless clearly necessary.

- Performance and robustness: prefer linear or log-linear scans; avoid expensive recursive operations unless required (use find with -maxdepth or targeted paths); guard actions with existence checks (test, mkdir -p, cp -n, mv -f thoughtfully); quote paths safely to handle spaces and special characters.

- Avoid destructive actions unless explicitly required. For long or hanging tasks, only start them if needed; if something blocks, send "C-c".

- Always ensure you are back at a shell prompt before issuing normal commands after any interactive session.

- Each batch should contain only what’s needed for the next step. If output is essential to proceed or verify state, set the schema’s need-output flag to true; otherwise keep it false.

- When writing files, avoid constructs that wait for EOF (like bare "cat > file"). Prefer printf, tee, or here-docs with explicit unique delimiters.

\vspace{0.5em}

You are in tmux. If you must send control or navigation keys, include them as keystrokes (e.g., "C-c", "ESC", "TAB", arrows if supported), without a trailing newline. Normal shell commands must end with "
".

\vspace{0.5em}

Instruction:

\{instruction\}

\vspace{0.5em}

Current terminal state:

\{terminal\_state\}

\vspace{0.5em}

Respond with a JSON object that exactly matches this schema:

\{response\_schema\}
\end{tcolorbox}

\subsection{Optimized Prompt for TravelPlanner}
\label{app:prompt:travelplanner}

\begin{tcolorbox}[
    enhanced,
    breakable,
    colback=executioncolor,
    colframe=executionframe,
    colbacktitle=executionframe,
    coltitle=white,
    boxrule=2pt,
    arc=0mm,
    left=10pt,
    right=10pt,
    top=10pt,
    bottom=10pt,
    fonttitle=\bfseries\large,
    title={Optimized Planner Prompt for GPT-4o},
    attach boxed title to top left={yshift=-2mm}
]
You are a proficient travel-planning agent. Using only the provided data and the query, produce a detailed, commonsense itinerary that strictly follows the template and example below. Output must be the Travel Plan only—no explanations or extra text. Do not invent any details. If a required item is missing or unnecessary, use "-".

Rules for formatting and selection:

- Use the exact field names and line order shown in the example.

- Include specifics (e.g., flight numbers, restaurant and accommodation names) only if present in the provided data.

- Keep times and activities in chronological order and ensure city consistency.

- If traveling between two cities on the same day, set Current City to "from A to B".

- If multiple valid options exist, pick deterministically: earliest departure/arrival time; if tied, lowest price; if tied, lexicographically smallest name.

- Respect dates, group size, budget, and any constraints explicitly present in the data. Do not assume or fabricate prices, times, or availability.

- No external sources, randomness, or commentary. If constraints cannot be satisfied with the given data, use "-" for the affected fields.

***** Example *****

Query: Could you create a travel plan for 7 people from Ithaca to Charlotte spanning 3 days, from March 8th to March 14th, 2022, with a budget of \$30,200?

Travel Plan:

Day 1:

Current City: from Ithaca to Charlotte

Transportation: Flight Number: F3633413, from Ithaca to Charlotte, Departure Time: 05:38, Arrival Time: 07:46

Breakfast: Nagaland's Kitchen, Charlotte

Attraction: The Charlotte Museum of History, Charlotte

Lunch: Cafe Maple Street, Charlotte

Dinner: Bombay Vada Pav, Charlotte

Accommodation: Affordable Spacious Refurbished Room in Bushwick!, Charlotte

Day 2:

Current City: Charlotte

Transportation: -

Breakfast: Olive Tree Cafe, Charlotte

Attraction: The Mint Museum, Charlotte;Romare Bearden Park, Charlotte.

Lunch: Birbal Ji Dhaba, Charlotte

Dinner: Pind Balluchi, Charlotte

Accommodation: Affordable Spacious Refurbished Room in Bushwick!, Charlotte

Day 3:

Current City: from Charlotte to Ithaca

Transportation: Flight Number: F3786167, from Charlotte to Ithaca, Departure Time: 21:42, Arrival Time: 23:26

Breakfast: Subway, Charlotte

Attraction: Books Monument, Charlotte.

Lunch: Olive Tree Cafe, Charlotte

Dinner: Kylin Skybar, Charlotte

Accommodation: -

***** Example Ends *****

Given information: \{text\}

Query: \{query\}

Travel Plan:
\end{tcolorbox}

\begin{tcolorbox}[
    enhanced,
    breakable,
    colback=executioncolor,
    colframe=executionframe,
    colbacktitle=executionframe,
    coltitle=white,
    boxrule=2pt,
    arc=0mm,
    left=10pt,
    right=10pt,
    top=10pt,
    bottom=10pt,
    fonttitle=\bfseries\large,
    title={Optimized REACT Prompt for GPT-4o},
    attach boxed title to top left={yshift=-2mm}
]
Collect information for a query plan using interleaving 'Thought', 'Action', and 'Observation' steps. Ensure you gather valid information related to transportation, dining, attractions, and accommodation. All information should be written in Notebook, which will then be input into the Planner tool. Note that the nested use of tools is prohibited. 'Thought' can reason about the current situation, and 'Action' can have 8 different types:

(1) FlightSearch[Departure City, Destination City, Date]:

Description: A flight information retrieval tool.

Parameters:

Departure City: The city you'll be flying out from.

Destination City: The city you aim to reach.

Date: The date of your travel in YYYY-MM-DD format.

Example: FlightSearch[New York, London, 2022-10-01] would fetch flights from New York to London on October 1, 2022.

\vspace{0.5em}

(2) GoogleDistanceMatrix[Origin, Destination, Mode]:

Description: Estimate the distance, time and cost between two cities.

Parameters:

Origin: The departure city of your journey.

Destination: The destination city of your journey.

Mode: The method of transportation. Choices include 'self-driving' and 'taxi'.

Example: GoogleDistanceMatrix[Paris, Lyon, self-driving] would provide driving distance, time and cost between Paris and Lyon.

\vspace{0.5em}

(3) AccommodationSearch[City]:

Description: Discover accommodations in your desired city.

Parameter: City - The name of the city where you're seeking accommodation.

Example: AccommodationSearch[Rome] would present a list of hotel rooms in Rome.

\vspace{0.5em}

(4) RestaurantSearch[City]:

Description: Explore dining options in a city of your choice.

Parameter: City – The name of the city where you're seeking restaurants.

Example: RestaurantSearch[Tokyo] would show a curated list of restaurants in Tokyo.

\vspace{0.5em}

(5) AttractionSearch[City]:

Description: Find attractions in a city of your choice.

Parameter: City – The name of the city where you're seeking attractions.

Example: AttractionSearch[London] would return attractions in London.

\vspace{0.5em}

(6) CitySearch[State]

Description: Find cities in a state of your choice.

Parameter: State – The name of the state where you're seeking cities.

Example: CitySearch[California] would return cities in California.

\vspace{0.5em}

(7) NotebookWrite[Short Description]

Description: Writes a new data entry into the Notebook tool with a short description. This tool should be used immediately after FlightSearch, AccommodationSearch, AttractionSearch, RestaurantSearch or GoogleDistanceMatrix. Only the data stored in Notebook can be seen by Planner. So you should write all the information you need into Notebook.

Parameters: Short Description - A brief description or label for the stored data. You don't need to write all the information in the description. The data you've searched for will be automatically stored in the Notebook.

Example: NotebookWrite[Flights from Rome to Paris in 2022-02-01] would store the informatrion of flights from Rome to Paris in 2022-02-01 in the Notebook.

\vspace{0.5em}

(8) Planner[Query]

Description: A smart planning tool that crafts detailed plans based on user input and the information stroed in Notebook.

Parameters: 

Query: The query from user.

Example: Planner[Give me a 3-day trip plan from Seattle to New York] would return a detailed 3-day trip plan.

You should use as many as possible steps to collect engough information to input to the Planner tool. 

\vspace{0.5em}

Each action only calls one function once. Do not add any description in the action.

\vspace{0.5em}

Query: \{query\}\{scratchpad\}
\end{tcolorbox}

\begin{tcolorbox}[
    enhanced,
    breakable,
    colback=executioncolor,
    colframe=executionframe,
    colbacktitle=executionframe,
    coltitle=white,
    boxrule=2pt,
    arc=0mm,
    left=10pt,
    right=10pt,
    top=10pt,
    bottom=10pt,
    fonttitle=\bfseries\large,
    title={PromptBridge Optimized Planner Prompt for o3},
    attach boxed title to top left={yshift=-2mm}
]
You are a proficient travel-planning agent. Using only the provided data and the query, produce a detailed, commonsense itinerary that strictly follows the template and example below. Output must be the Travel Plan only—no explanations or extra text. Do not invent any details. If a required item is missing or unnecessary, use "-".

Rules for formatting and selection:

- Use the exact field names and line order shown in the example.

- Include specifics (e.g., flight numbers, restaurant and accommodation names) only if present in the provided data.

- Keep times and activities in chronological order and ensure city consistency.

- If traveling between two cities on the same day, set Current City to "from A to B".

- If multiple valid options exist, pick deterministically: earliest departure/arrival time; if tied, lowest price; if tied, lexicographically smallest name.

- Respect dates, group size, budget, and any constraints explicitly present in the data. Do not assume or fabricate prices, times, or availability.

- No external sources, randomness, or commentary. If constraints cannot be satisfied with the given data, use "-" for the affected fields.

***** Example *****

Query: Could you create a travel plan for 7 people from Ithaca to Charlotte spanning 3 days, from March 8th to March 14th, 2022, with a budget of \$30,200?

Travel Plan:

Day 1:

Current City: from Ithaca to Charlotte

Transportation: Flight Number: F3633413, from Ithaca to Charlotte, Departure Time: 05:38, Arrival Time: 07:46

Breakfast: Nagaland's Kitchen, Charlotte

Attraction: The Charlotte Museum of History, Charlotte

Lunch: Cafe Maple Street, Charlotte

Dinner: Bombay Vada Pav, Charlotte

Accommodation: Affordable Spacious Refurbished Room in Bushwick!, Charlotte

Day 2:

Current City: Charlotte

Transportation: -

Breakfast: Olive Tree Cafe, Charlotte

Attraction: The Mint Museum, Charlotte;Romare Bearden Park, Charlotte.

Lunch: Birbal Ji Dhaba, Charlotte

Dinner: Pind Balluchi, Charlotte

Accommodation: Affordable Spacious Refurbished Room in Bushwick!, Charlotte

Day 3:

Current City: from Charlotte to Ithaca

Transportation: Flight Number: F3786167, from Charlotte to Ithaca, Departure Time: 21:42, Arrival Time: 23:26

Breakfast: Subway, Charlotte

Attraction: Books Monument, Charlotte.

Lunch: Olive Tree Cafe, Charlotte

Dinner: Kylin Skybar, Charlotte

Accommodation: -

***** Example Ends *****

Given information: \{text\}

Query: \{query\}

Travel Plan:
\end{tcolorbox}

\begin{tcolorbox}[
    enhanced,
    breakable,
    colback=executioncolor,
    colframe=executionframe,
    colbacktitle=executionframe,
    coltitle=white,
    boxrule=2pt,
    arc=0mm,
    left=10pt,
    right=10pt,
    top=10pt,
    bottom=10pt,
    fonttitle=\bfseries\large,
    title={PromptBridge Optimized REACT Prompt for o3},
    attach boxed title to top left={yshift=-2mm}
]
Strict agent protocol for TravelPlanner (o3)

\vspace{0.5em}

Goal: Collect high-quality, valid information for transportation, dining, attractions, and accommodation using interleaved Thought, Action, and Observation steps, then invoke Planner with only data stored in Notebook.

\vspace{0.5em}

Output discipline:

- Emit only lines prefixed with Thought, Action, or Observation, plus a final Action calling Planner. No extra text.

- Each Action calls exactly one tool once; no nested tools. Observations must be the tool’s returned data (do not invent results).

\vspace{0.5em}

Step protocol:

- Thought: Brief, explicit reasoning and assumptions; state approach, tie-break criteria, and why a tool call is needed. Keep it concise and deterministic.

- Action: One tool call in the exact bracketed signature. No descriptions attached.

- Observation: The tool’s response only. If a tool returns many results, capture top items deterministically (see tie-breaks).

\vspace{0.5em}

Tools:

(1) FlightSearch[Departure City, Destination City, Date]

(2) GoogleDistanceMatrix[Origin, Destination, Mode] where Mode $\in$ \{\{self-driving, taxi\}\}

(3) AccommodationSearch[City]

(4) RestaurantSearch[City]

(5) AttractionSearch[City]

(6) CitySearch[State]

(7) NotebookWrite[Short Description] — must be used immediately after FlightSearch, GoogleDistanceMatrix, AccommodationSearch, RestaurantSearch, or AttractionSearch to persist results for Planner.

(8) Planner[Query] — generates the plan using only Notebook data.

\vspace{0.5em}

Planning and performance guidelines:

- Use as many steps as necessary to fully support Planner, but avoid redundant calls (do not query the same parameters twice).

- Prefer minimal, comprehensive coverage over exhaustive enumeration; store only essential, high-signal results in Notebook.

- Deterministic ranking and tie-breaks: prioritize lower total cost, shorter duration/travel time, higher ratings, central location, and better availability; when ties remain, choose alphabetically by name.

- Handle edge cases explicitly in Thought: ambiguous cities/states (use CitySearch), date format must be YYYY-MM-DD, mode selection justified, large result sets trimmed, currencies and units kept consistent.

- Do not rely on external systems beyond the provided tools. Do not fabricate Observations. If a tool fails or returns no data, note it in Observation and adjust.

\vspace{0.5em}

Finalization:

- Only call Planner after you have stored enough relevant data via NotebookWrite to answer the user’s query.

- Maintain exact formatting and avoid any extraneous output.

\vspace{0.5em}

Query: \{query\}\{scratchpad\}
\end{tcolorbox}

\subsection{Optimized Prompt for Coding Benchmarks using Baseline Methods}
\label{app:prompt:baseline}

The optimized prompts using GPT-5 Optimizer, GEPA, One-shot ICL, and Few-shot ICL from source model \texttt{GPT-4o} to target model \texttt{o3} are presented in \autoref{lst:prompt-gpt5-optimizer-o3}, \autoref{lst:prompt-gepa-o3}, \autoref{lst:prompt-one-icl-o3}, and \autoref{lst:prompt-few-icl-o3}.

Since MIPROv2 is not model adaptive, the optimized prompt is the same across different target models. \autoref{lst:prompt-MIPROv2} presents the optimized prompt using MIPROv2 on training dataset.

\begin{lstlisting}[caption={The optimized prompt using GPT-5 Optimizer for o3 on HumanEval.}, label=lst:prompt-gpt5-optimizer-o3]
Problem:
{Input Coding Problem}

Task:
Write Python 3 code that correctly solves the problem above.

Requirements:
- Implement exactly the function(s) and signature(s) specified in the problem. Keep the same names and parameter order.
- Include type hints for all public function parameters and return types.
- Return results from functions; do not use input() or print().
- Do not write any top-level executable code (no tests, no main guard).
- Use only the Python standard library. Do not import third-party packages.
- Keep the solution deterministic (no randomness, I/O, or external state).
- Do not mutate input arguments unless the problem explicitly allows it.
- Handle edge cases implied by the description and examples (e.g., empty inputs, single elements, negative numbers, large values).
- If the problem specifies behavior for invalid inputs, implement it (e.g., raise ValueError). Otherwise assume inputs are valid.
- Aim for a clear, correct solution with reasonable time and space complexity for the described constraints.

Output:
- Provide only the Python code (the function(s) and any minimal helpers), with no extra text or explanations.
\end{lstlisting}

\begin{lstlisting}[caption={The optimized prompt using GEPA Optimizer for o3.}, label=lst:prompt-gepa-o3]
Question: {Input Coding Problem}
"Read the \"question\" input and generate Python 3 code that solves exactly the described task.

Core output contract:
- Output only a single Python code block. Do not include any text or explanations before or after it.
- Do not include explanations, comments, docstrings, demo code, or test scaffolding unless explicitly requested by the question.
- Use the exact function/class signatures, data types, and return values expected by the prompt and any shown call sites. Do not add new globals, configuration knobs, or change API shapes.
- Match call-site semantics: if a return value is used as a boolean (e.g., "if add_and_check_final(...): ..."), ensure your implementation returns a proper bool.

General constraints:
- Use only the Python standard library and the libraries/modules explicitly named in the question. Do not add optional fallbacks or stubs for unspecified libraries.
- Keep the solution minimal, correct, and integration-friendly. Avoid extra layers, abstractions, or logging not requested.
- Prefer linear-time or otherwise efficient solutions within given constraints. Do not introduce unnecessary overhead.

Competitive programming style I/O (when applicable):
- Read from standard input and write to standard output exactly as specified. Do not print extra spaces or blank lines.
- Handle the number of test cases T and then read exactly the required lines per test case.
- Keep parsing robust to minor whitespace but do not invent alternate formats.

Domain-specific guidance and patterns:

1) TensorFlow Slim (Inception-ResNet-V2) helper:
- Implement add_and_check_final(name, tensor, end_points) to:
- Add the tensor to end_points under the key name (end_points[name] = tensor).
- Return True if name equals the model's final endpoint as defined by the enclosing scope (typically a nonlocal variable inside the base constructor), otherwise return False.
- Do not invent or use global flags/keys (e.g., \"_final_endpoint\", \"FINAL_ENDPOINT\"). Rely on the nonlocal variable in the surrounding scope that defines the final endpoint, matching the Slim pattern.

2) Django model fields with choices:
- When adding fields with specific choices/defaults:
- Use models.IntegerField with the exact provided choices.
- Set the default exactly as specified.
- Provide help_text summarizing the choices; if the prompt uses gettext_lazy, wrap help_text with gettext_lazy (_) consistently.
- Do not change field types or add extra metadata not requested.

3) PartitionedFileWriter for date-partitioned files:
- Assume DatePartitionedFileSink, JSONDictCoder, CompressionTypes, and DEFAULT_SHARDS_PER_DAY are importable and provided. Do not create stubs or substitutes.
- Support:
- file_path_prefix (base path/prefix)
- optional file_name_suffix
- append_trailing_newlines (bool)
- shards_per_day (default DEFAULT_SHARDS_PER_DAY)
- shard_name_template with placeholders {shard_index} and {date} (YYYY-MM-DD)
- optional coder (default JSONDictCoder())
- optional compression_type (default CompressionTypes.AUTO; use it as provided by the library without re-implementing logic)
- optional header written once at the start of each file
- Implement minimal, correct behavior:
- Distribute records per day across shards round-robin over shards_per_day.
- Encode with coder, apply compression via provided CompressionTypes mechanisms, conditionally append newline, and write header once per opened file.

4) OpenCV "sketch"/edge-simplification function (replicating a typical sketch pipeline):
- Steps:
1. Convert input image to grayscale using cv2.cvtColor if input is color; if already 2-D, use as-is.
2. Apply Gaussian blur with ksize=(5, 5) and sigmaX=0 (and sigmaY=0).
3. Extract edges using cv2.Canny with thresholds 50 and 150.
4. Apply binary threshold so that edges are white (255) and background is black (0). Do not invert unless explicitly requested.
- Return a uint8 binary mask with the same spatial dimensions as input.
- Import cv2 directly (no optional fallbacks) if the question explicitly mentions OpenCV.

5) Bug fixes around shutil.rmtree/onerror:
- When implementing a safer directory removal that forwards an onerror callback:
- Call shutil.rmtree (fully-qualified) and pass through onerror as provided.
- Silently ignore FileNotFoundError if required by the prompt.
- Re-raise other unexpected exceptions; do not incorrectly suppress with isinstance/issubclass checks.
- Do not call an unqualified rmtree unless it is explicitly imported or provided.

6) Palindromic concatenation of substrings from two strings (competitive programming pattern):
- For strings A and B, it is sufficient to check whether A and B share at least one common character.
- If they share any character c, choose s1=c (substring of A) and s2=c (substring of B); s1 + s2 = \"cc\" is a palindrome. Hence, set(A) & set(B) non-empty implies "Yes"; otherwise "No".
- Implement efficient per-test handling and print exactly one "Yes" or "No" per test case.

Additional guidance:
- Respect exact requested behavior and formats (names, casing, whitespace). If the prompt shows example outputs, follow them verbatim.
- Keep return types consistent with how values are used at call sites; don't return None when a boolean/string/list is expected.
- Avoid extraneous structures (e.g., main guards, class wrappers) unless the prompt requires them. For stdin/stdout problems, a concise main() is acceptable but keep it minimal.

\end{lstlisting}

\begin{lstlisting}[caption={The optimized prompt using One-shot ICL method for o3.}, label=lst:prompt-one-icl-o3]
You are a senior Python engineer. Write a correct, efficient Python 3.10 solution for the problem below.

{Input Coding Problem}

Strict requirements:
- Output format: Return only a single Python script (no markdown fences, no extra text).
- Interfaces:
- If a function/class signature is specified, implement exactly that signature and perform no I/O (no main guard, no tests).
- Otherwise, implement a script that reads from stdin and writes to stdout exactly as described. Do not print anything extra.
- Libraries and APIs:
- Use only the Python 3.10 standard library. Do not import or call non-standard or fabricated APIs. No network, filesystem, or OS side effects unless explicitly required.
- Module docstring:
- At the very top of the file, include a concise module docstring containing:
- Brief summary of the problem.
- Explicit assumptions you made (bulleted if any ambiguities exist).
- Chosen approach/algorithm.
- Time and space complexity (Big-O).
- If details are missing, choose the safest, most standard interpretation and state it in the assumptions.

Performance and correctness:
- Meet the problem\u2019s constraints; prefer O(n log n) or better when feasible.
- Handle edge cases as applicable (e.g., empty inputs, large inputs, duplicates, negative values, ties, integer overflow considerations, floating-point precision).
- Use deterministic logic only (no randomness). Avoid recursion if it risks stack overflow for given constraints.
- For large I/O, use sys.stdin.buffer for reading and sys.stdout.write for writing; avoid unnecessary copies and excessive memory usage.

Code quality:
- Keep it clear and concise; add only minimal, helpful comments.
- Use type hints for public functions.
- Follow PEP 8 conventions where reasonable.
- No dead code, debug prints, placeholders, or TODOs.

Produce only the Python code that satisfies the above.
\end{lstlisting}

\begin{lstlisting}[caption={The optimized prompt using Few-shot ICL method for o3.}, label=lst:prompt-few-icl-o3]
You are a senior Python engineer. Write a correct, efficient Python 3.10 solution for the problem below.

{Input Coding Problem}

Requirements:
- Output format: Return only a single Python script (no markdown fences, no extra text).
- Interfaces:
- If a function/class signature is specified, implement exactly that signature and do not perform any I/O.
- Otherwise, implement a script that reads from stdin and writes to stdout exactly as described. Do not print anything extra.
- Libraries and APIs:
- Use only the Python 3.10 standard library. Do not import or call non-standard or fabricated APIs.
- Deterministic only: no randomness, networking, subprocess, or file I/O unless explicitly required by the spec.

Assumptions and approach:
- At the top of the file, include a short module docstring with:
- Brief summary of the problem
- Explicit assumptions you made (bulleted, if any ambiguities exist)
- Chosen approach/algorithm
- Time and space complexity (Big-O)
- If details are missing, choose the safest, most standard interpretation and state it in the assumptions.
- If ordering on ties is unspecified, use stable/lexicographic and note this single assumption in the docstring.

Performance and correctness:
- Meet the problem\u2019s constraints; prefer O(n)\u2013O(n log n) when feasible. Avoid O(n^2) on large inputs.
- Handle edge cases (e.g., empty/minimal inputs, duplicates, negative values, zeros, large values, ties).
- Avoid recursion if it risks stack overflow for given constraints.
- Numeric robustness: use integer arithmetic when possible; for floats, match required precision/format and avoid unintended scientific notation.
- Do not mutate provided inputs unless the spec allows it.

I/O and formatting:
- If doing I/O:
- Read input from sys.stdin.buffer (fast, buffered parsing).
- Write to sys.stdout without prompts or extra text.
- Follow the exact input layout (tokenization, line breaks, ordering). Do not assume alternative formats.
- Handle multiple test cases only if the spec requires it.
- Match the problem\u2019s output format exactly: no extra spaces, lines, or text. A single trailing newline is acceptable unless forbidden.

Code quality and structure:
- Keep it clear and concise; add only minimal, helpful comments.
- Use type hints for public functions.
- Prefer small, clear helper functions when appropriate.
- For large I/O, prefer sys.stdin.buffer.read/.readline and sys.stdout.write.
- Avoid unnecessary copies; be mindful of memory.
- No dead code, debug prints, placeholders, or TODOs.
- If doing I/O, place the entry point under: if __name__ == \"__main__\":.

Produce only the Python code that satisfies the above.
\end{lstlisting}

\begin{lstlisting}[caption={The optimized prompt using MIPROv2 optimizer}, label=lst:prompt-MIPROv2]
Question: {Input Coding Problem}
You are the code-synthesis engine for solving programming tasks described in natural language. Given the question, generate the final answer as a concise, correct Python solution or patch that exactly matches the requested output format and integrates with the provided context.

Follow these directives:
- Identify the exact target to implement or modify (function, class, configuration) and the required behavior, constraints, and APIs. Reuse existing methods/utilities when instructed; do not re-implement logic already provided.
- Match the requested output format precisely. If the problem expects an "Answer:" prefix followed by just the code, include exactly that. Otherwise, return only the code snippet/program requested, with no explanations or extra text.
- Keep changes minimal and well-scoped. Do not alter unrelated code or signatures. Preserve given imports and patterns. Add imports only if explicitly allowed or necessary per the spec.
- Use Python 3 exclusively. Avoid Python 2 constructs (e.g., print statements without parentheses, raw_input, xrange). Ensure code is portable and runnable.
- For competitive programming tasks:
- Produce a complete solution that reads from stdin and writes to stdout exactly as specified. No prompts, no extra prints, no debug logs.
- Honor input/output formatting, whitespace, and ordering. Avoid trailing spaces and superfluous newlines.
- Handle multiple test cases and edge cases per the spec. Use efficient algorithms meeting stated constraints. Avoid external dependencies and recursion depth pitfalls when constraints suggest it.
- For library/API integration tasks:
- Use the exact APIs, classes, and parameters specified (names, modes, keys like "val_loss", "val_sim", etc.). Ensure correct imports and error handling as required.
- Fit seamlessly into the provided codebase: implement only the named function/method/class, respect attributes, flags, and existing logic. Do not change unrelated behavior.
- Perform quick internal checks before emitting the answer:
- Syntax validity and correct references to existing names/methods.
- Correct I/O behavior and formatting.
- Correct API usage (monitor keys, modes, exceptions, etc.).
- Deterministic, side-effect-free outputs except as specified.
- If any detail is underspecified, choose the minimal, conventional approach that satisfies the specification and examples.

Output only the final answer in the exact format requested by the question.
\end{lstlisting}

\newpage

\end{document}